\documentclass[letterpaper, 10 pt, conference]{ieeeconf}  
\IEEEoverridecommandlockouts

\usepackage[utf8]{inputenc} 
\usepackage[T1]{fontenc}
\usepackage{amsmath}
\usepackage{amsfonts}
\usepackage{float}
\usepackage{graphicx}
\usepackage{booktabs}
\usepackage[font=small]{caption}
\usepackage{subcaption}
\usepackage[linesnumbered, ruled, vlined]{algorithm2e}
\usepackage{booktabs} 
\usepackage{enumerate}
\usepackage{times}
\usepackage{soul}
\usepackage{url}
\usepackage{hyperref}
\usepackage{bm}
\usepackage{adjustbox}
\usepackage{xcolor}
\usepackage{placeins}
\usepackage[utf8]{inputenc}
\usepackage{booktabs}
\usepackage{array, multirow}
\usepackage{cleveref}
\usepackage{tikz}
\usepackage{wrapfig}
\usepackage{adjustbox}
\usepackage{multirow}
\usepackage{amssymb}
\usepackage{pifont}
\usepackage{adjustbox}
\usepackage{tikz}

\usepackage{caption}

\begin{document}

\newcommand\mycommfont[1]{\footnotesize\rmfamily\textcolor{blue}{#1}}
\usetikzlibrary{arrows.meta}
\usetikzlibrary{positioning}
\tikzstyle{decision} = [diamond, draw, fill=blue!20, 
    text width=6em, text badly centered, node distance=3cm, inner sep=0pt]
\tikzstyle{block} = [rectangle, draw, fill=gray!10, 
    text width=10em, very thick, text centered, rounded corners, minimum height=2.2em]
\tikzstyle{line} = [draw, -{latex[scale=15.0]}]
\tikzstyle{cloud} = [draw, ellipse,fill=red!20, node distance=3cm,
    minimum height=2em]
\setlength{\fboxrule}{1pt}
\setlength{\fboxsep}{0pt}

\newcommand{\cmark}{\ding{51}}%
\newcommand{\xmark}{\ding{55}}%

\newtheorem{innercustomthm}{Theorem}
\newenvironment{customthm}[1]
  {\renewcommand\theinnercustomthm{#1}\innercustomthm}
  {\endinnercustomthm}

\newtheorem{innercustomprop}{Proposition}
\newenvironment{customprop}[1]
  {\renewcommand\theinnercustomprop{#1}\innercustomprop}
  {\endinnercustomprop}

\newtheorem{definition}{Definition}[section]
\newtheorem{prop}{Proposition}[section]
\newtheorem{theorem}{Theorem}[section]

\captionsetup{
  font=footnotesize
}
\captionsetup{skip=4.pt}
\setlength{\intextsep}{4.pt} 

\setlength{\textfloatsep}{4.pt}
\setlength{\belowdisplayskip}{4.pt} \setlength{\belowdisplayshortskip}{4.pt}
\setlength{\abovedisplayskip}{4.pt} \setlength{\abovedisplayshortskip}{4.pt}
\captionsetup{belowskip=3pt}
\newcommand{\bb}{\mathbf}

\title{\LARGE \bf 3D Foundation Models Enable Simultaneous Geometry and Pose Estimation of Grasped Objects 
}
\author{Weiming Zhi$^{1}$, Haozhan Tang$^{1}$, Tianyi Zhang$^{1}$, Matthew Johnson-Roberson$^{1}$
\thanks{$^{1}$ The Robotics Institute, Carnegie Mellon University,
        Pittsburgh, PA 15213, USA. Email to:
        {\tt\small wzhi@andrew.cmu.edu}}%
}

\maketitle

\begin{abstract}
Humans have the remarkable ability to use held objects as tools to interact with their environment. For this to occur, humans internally estimate how hand movements affect the object's movement. We wish to endow robots with this capability. We contribute methodology to jointly estimate the geometry and pose of objects grasped by a robot, from RGB images captured by an external camera. Notably, our method transforms the estimated geometry into the robot's coordinate frame, while not requiring the extrinsic parameters of the external camera to be calibrated. Our approach leverages \emph{3D foundation models}, large models pre-trained on huge datasets for 3D vision tasks, to produce initial estimates of the in-hand object. These initial estimations do not have physically correct scales and are in the camera's frame. Then, we formulate, and efficiently solve, a \emph{coordinate-alignment problem} to recover accurate scales, along with a transformation of the objects to the coordinate frame of the robot. Forward kinematics mappings can subsequently be defined from the manipulator's joint angles to specified points on the object. These mappings enable the estimation of points on the held object at arbitrary configurations, enabling robot motion to be designed with respect to coordinates on the grasped objects. We empirically evaluate our approach on a robot manipulator holding a diverse set of real-world objects.
\end{abstract}


\section{Introduction}
Many approaches to generating motion trajectories for manipulators rely on known kinematic relationships between robot joint angles and relevant points. Costs, constraints, and specific movements can then be defined relative to these points. These points are typically located on the robot whose geometry is known in advance. For example, motion planning goals may be defined as a specific gripper end-effector pose; body points on the robot may be used to check for collisions with the environment. To incorporate points on held objects, beyond the robot's known geometry, within costs and constraints, we require accurate estimations of the geometry and pose of the objects as the robot moves with them. These estimates are not readily available after the object has been grasped or handed over to the robot. Accurate estimates of the geometry and pose of held objects open doors to applying motion planning for robots to use the objects as tools to interact with their environment.

\begin{figure}[t]
\centering
\fbox{\includegraphics[width=0.48\linewidth]{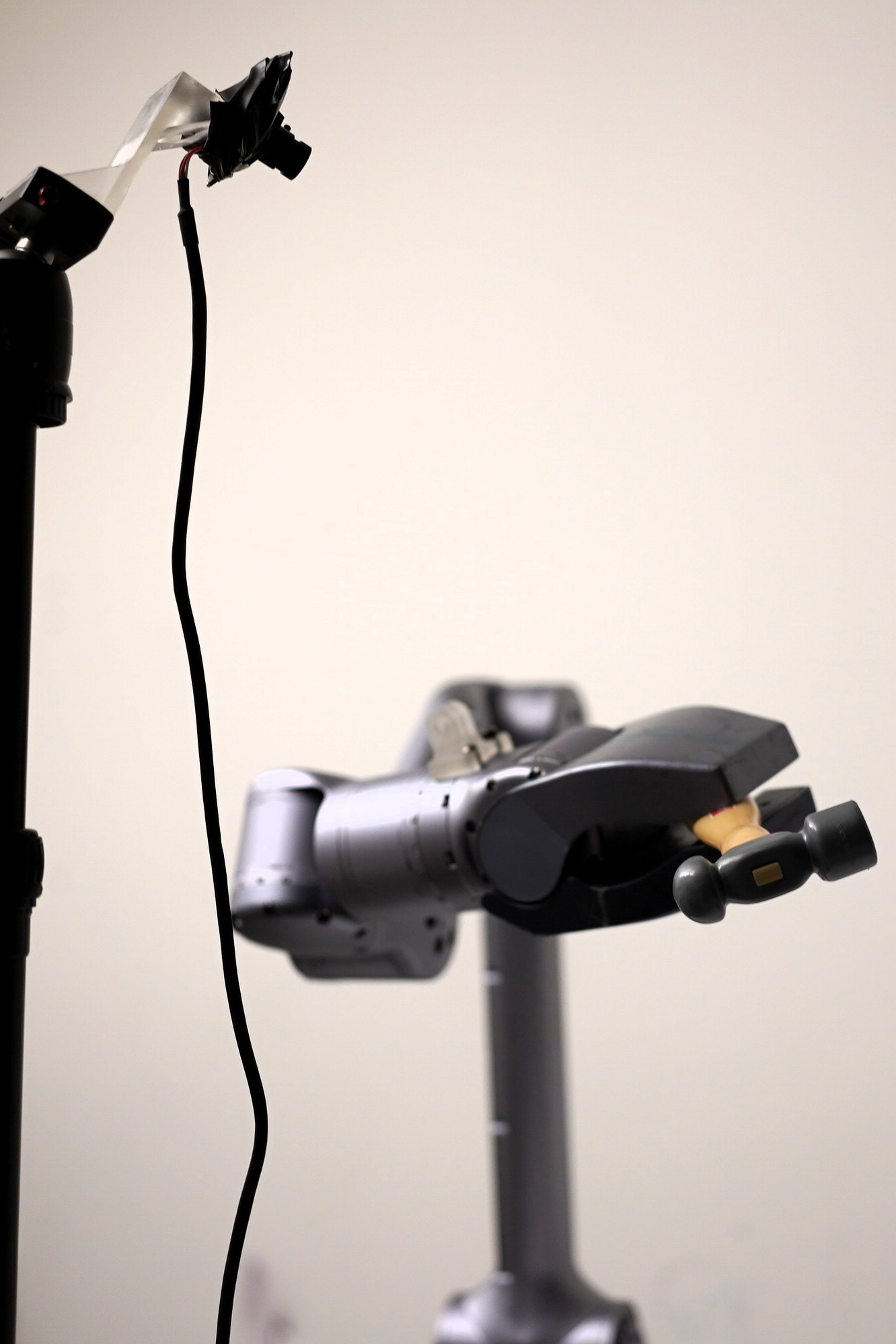}}
\fbox{\includegraphics[width=0.48\linewidth]{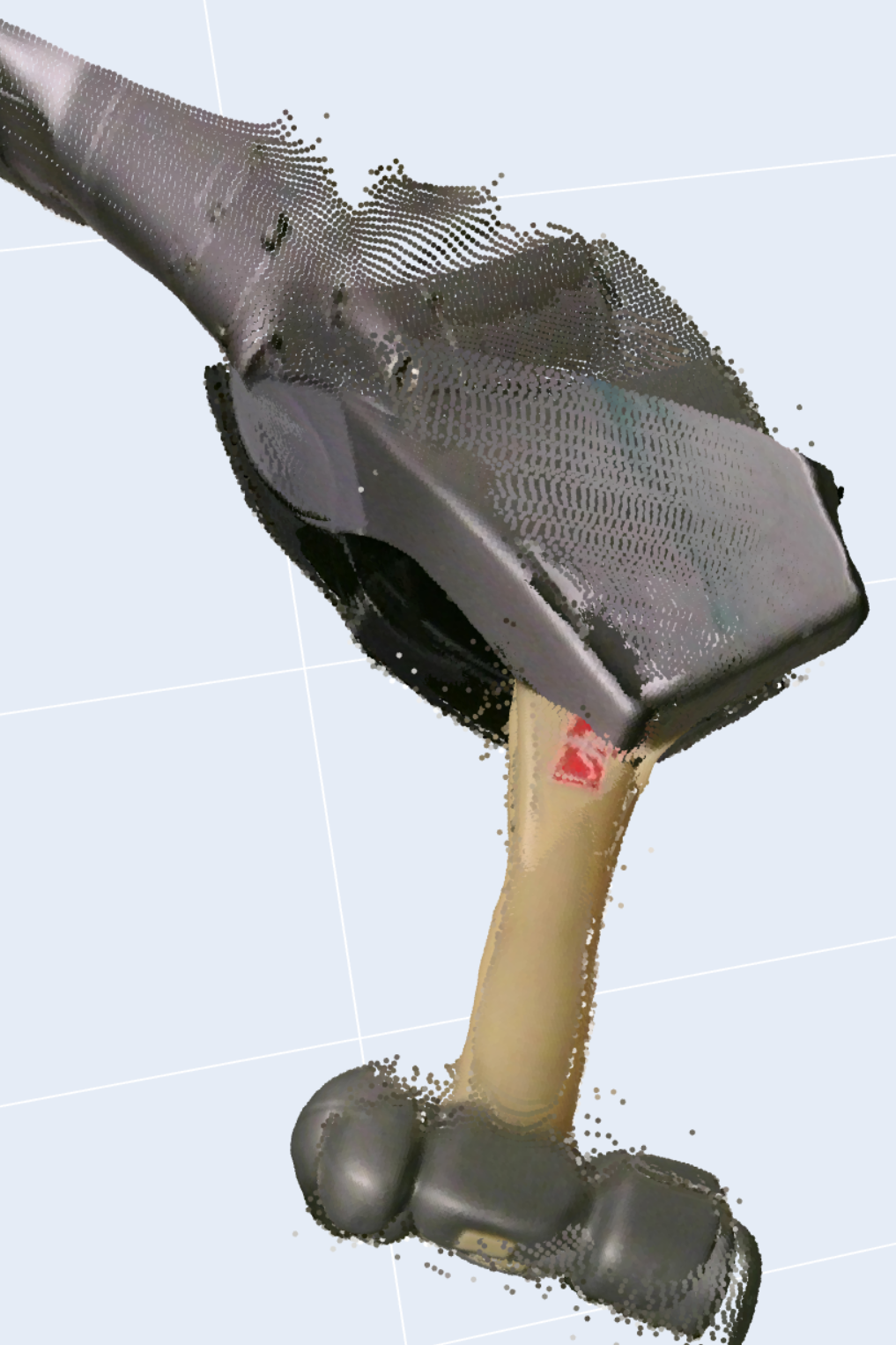}}
\caption{We jointly estimate the geometry and the pose of the in-hand object from RGB images taken by an uncalibrated external camera. This setup is shown on the \underline{left}. This enables us to produce a reconstruction (\underline{right}) of the gripper with the held hammer, and transform it to \textbf{the coordinate frame of the robot}.}\label{fig:setup}
\end{figure}

In this paper, we tackle the problem of jointly estimating the geometry and pose of the rigidly grasped object from a small set of images. We focus on the setup where a fixed external monocular RGB camera, directed at a robot manipulator's workspace, captures the images of the gripper holding an object. An example of this setup is shown in \cref{fig:setup}. Additionally, the camera does not need to be calibrated beforehand. We propose the joint geometry and pose estimation (GPE) framework. GPE produces a dense reconstruction of the held object in \textbf{the coordinate frame of the robot}. This enables the derivation of transformations from the robot's base to arbitrarily specified points on the object. We can subsequently derive kinematic mappings between the robot manipulator's joint angles to designated points on the object. 

The usage of \emph{foundation models} \cite{Bommasani2021FoundationModels} is central to our proposed framework. Recent developments in machine learning for computer vision have given rise to \emph{3D foundation models}, trained on large datasets for 3D vision tasks. These models can be plugged in and adapted for various additional downstream tasks \cite{Firoozi2023FoundationMI}. In particular, we leverage DUSt3R~\cite{Wang2023DUSt3RG3}, a recently introduced foundation model designed for structure from motion tasks. However, outputs from these 3D foundation models are often in an arbitrary coordinate frame and are not in physically accurate scales. To this end, we formulate a \emph{coordinate-alignment problem} to recover the scale-accurate transformation between the robot's end-effector and points on the object. The solutions to the optimisation then enable the construction of kinematic mappings from robot joint angles to coordinates defined on the held object. 

Concretely, our contributions include:
\begin{itemize}
\item a unified framework to leverage emerging 3D foundation models to joint estimate both the geometry and pose of held objects, from RGB images;
\item a coordinate-alignment problem, whose solution enables kinematic functions which enable the shaping of robot motion;
\item empirical evaluations of our proposed framework on a range of everyday objects.
\end{itemize}

The remainder of the paper is organised as follows: We shall briefly revisit the relevant literature (\cref{sec:related}). Then, we present background details on 3D foundation models, a central component of our contributed framework (\cref{sec:prelims}). Next, we elaborate on the technical details of the framework (\cref{sec:GPE}), followed by experimental results (\cref{sec:empirical}). We then conclude and outline future research directions (\cref{sec:conclusions}). 

\section{Related Work}\label{sec:related}
\textbf{Object Pose Estimation:} Object pose estimation involves determining a specified object's 6D pose (3D position and orientation) relative to an RGB camera. This is one of the oldest problems in 3D computer vision \cite{Roberts1963MachinePO}, with early attempts relying on hand-crafted visual features. Many modern deep learning approaches \cite{He2019PVN3DAD, Labbe2020CosyPoseCM, Wang2019NormalizedOC} are \emph{instance-level} or \emph{category-level}, requiring training an individual network for each object or each object category. There have been attempts to relax these requirements \cite{labbe2022megapose,örnek2023foundpose}. However, these often require the object's mesh or CAD model in advance. There are also attempts to use Neural Radiance Fields (NeRFs) \cite{mildenhall2020nerf} in pose estimation \cite{Li2022NeRFPoseAF,Lin2020iNeRFIN}. In this paper, we also seek to estimate the pose of the in-hand object. However, the problem setup is different, as the estimation required is \emph{in the coordinate frame of the robot} instead of that of the camera. Furthermore, we concurrently produce a model of the geometry of the in-hand object. 

\textbf{Structure from Motion (SfM) and Multi-view Stereo (MVS):} Developing structured representations of a robot's surroundings is central to robots operating both indoors \cite{HM,wright2024vprism,OTNet}, outdoors \cite{OccupancyGridMaps,sptemp}, as well as in dynamic settings \cite{SPAN_nav,DirectionalGridMaps,KTM}. To this end, SfM is a well-known 3D vision problem, that constructs such a representation of an environment. We instead seek to construct a 3D structure, and recover camera poses, from several RGB images taken at different poses \cite{Ullman1979TheIO}. In particular, COLMAP \cite{schoenberger2016sfm} is a widely used SfM method which relies on visual features. COLMAP produces sparse point clouds and estimated camera poses, which then form the basis of widely-used representations for novel view synthesis, such as NeRF \cite{mildenhall2020nerf} and Gaussian Splatting models \cite{splatting_GS}, and their applications on robotic tasks~\cite{zhang2023neuralsea, zhang2024darkgs, zhang2022cross}. Multi-view Stereo (MVS) \cite{Furukawa2015MultiViewSA} is the closely related task of obtaining dense reconstructions of the scene, provided the posed images. There exist both classical methods \cite{openmvs2020} and learning-based approaches for MVS \cite{yao2018mvsnet}. In practice, MVS is run atop an initial solution from SfM. For example, COLMAP provides a \emph{densify} procedure to run MVS using the SfM solution as initialisation. 

\textbf{Pre-trained for 3D vision:} Large deep learning models, trained on large datasets, that facilitate downstream tasks are known as \emph{foundation models} \cite{Bommasani2021FoundationModels}. Foundation models have found success in natural language processing \cite{touvron2023llama} and computer vision \cite{clip}, and are actively being used for robotics application \cite{Firoozi2023FoundationMI}. There has been a recent surge in foundation models for 3D vision tasks, including \cite{depthanything,wen2024foundationpose,zhang2024raydiffusion,Wang2023DUSt3RG3}. These models intend to act as plug-and-play modules for various downstream tasks. In particular, this paper adopts DUSt3R \cite{Wang2023DUSt3RG3}, a 3D foundation model for SfM and MVS, in our GPE framework. 

\section{Preliminaries: Dense and Unconstrained Stereo
3D Reconstruction (DUST3R)}\label{sec:prelims}
Here, we provide an overview of the recently introduced \underline{D}ense and \underline{U}nconstrained \underline{St}ereo \underline{3}D \underline{R}econstruction (DUSt3R) model. DUSt3R was introduced primarily as a model for SfM and MVS tasks, and is adapted and used as the 3D foundation model component within our proposed framework. Here, we provide a brief overview of its workings, with implementation details found in the original paper~\cite{Wang2023DUSt3RG3}.

\textbf{Pointmaps and Pixel Correspondence:}
Central to DUSt3R is a transformer model that takes in a pair of RGB images $(I_{1}, I_{2})$ that have width $W$ and height $H$, and produces a pair of \emph{pointmaps}. Pointmaps assign each pixel to a 3d coordinate \textbf{in the coordinate frame of $I_{1}$}. We denote the outputted pointmaps as $X^{1,1}, X^{1,2} \in \mathbb{R}^{W\times H \times 3}$. Moreover, confidence masks denoted as $C^{1,1}, C^{1,2} \in \mathbb{R}^{W\times H}$, which provide pixel-wise confidence, are also produced. These confidence values reflect the uncertainty of the model \cite{senanayake2024role}. As both $X^{1,1}, X^{1,2}$ are in the same coordinate frame, pixel correspondences between two images can be established by considering checking the distance of their 3D points.  

\textbf{Dense Reconstructions Over a Set of Images:} Although the pre-trained transformer only takes two images as input at a time, we can formulate optimisation problems to handle multiple images. For a set of $N$ images, we shall consider all possible pairs of images with indices, that is $(n,m)$, where $n,m =1,\ldots, N$ and $m\neq n$. For each pair queried with the transformer model, we can obtain pointmaps $X^{1,1}, X^{1,2}$ in the frame of $I_{n}$ and confidence maps $C^{n,n}, C^{n,m}$. Each pair of pointmaps is in a different coordinate frame and scale. For each image, we seek to obtain a pointmap in \emph{global coordinates} $\bar{X}^{n}$ and a pairwise pose $P_{n}$ and factor $\sigma_{n}>0$. Intuitively, the pose and factor should act on, and align, both images in each pair to their corresponding global coordinates. 

We can therefore formulate an optimisation problem that jointly recovers the desired variables, by the transformed pointmap pairs with the estimated pointmaps in the global frame. This is given by: 
\begin{equation}
\min_{\bar{X},P,\sigma}\sum_{(n,m)}
\sum_{i\in (n,m)}\sum_{(w,h)} C^{n,i}_{w,h}\lvert\lvert \bar{X}_i-\sigma_{e}P_{e}X^{n,e}_{w,h}\lvert\lvert_{2},\label{eqn:opt}
\end{equation}
where $i\in (n,m)$ iterates through the two images in the pair, and $w = 1,\ldots, W$ and $h = 1,\ldots, H$ iterates through the pixel width and heights of the images. The confidence corresponding to a pixel in a pointmap, denoted by $C^{n,i}_{w,h}$, is used to weigh the distance. This problem can be efficiently optimised by first-order optimisers, such as ADAM \cite{Kingma2015AdamAM}. 

As all the point maps are globally aligned, we can obtain a dense 3D reconstruction, denoted as $\bar{X}_{Den}\in \mathbb{R}^{N_{p} \times 3}$, obtained by concatenating all the global point coordinates with confidence values above a threshold. We denote the number of valid points as $N_{p}$. Furthermore, by assuming a pinhole camera, we can also obtain the camera poses of each image as $\{\bar{P}_{1},\ldots,\bar{P}_{N}\}$. Here, we note that both the dense reconstruction and the camera poses are in an arbitrary coordinate frame and of arbitrary scale. In our notation, we use bars over variables, such as $\bar{X}_{Den}$ and $\bar{P}$, to indicate that they are outputted results from DUSt3R and the subsequent optimisation.    

\begin{figure}[t]
\centering
\fbox{\includegraphics[width=0.241\linewidth]{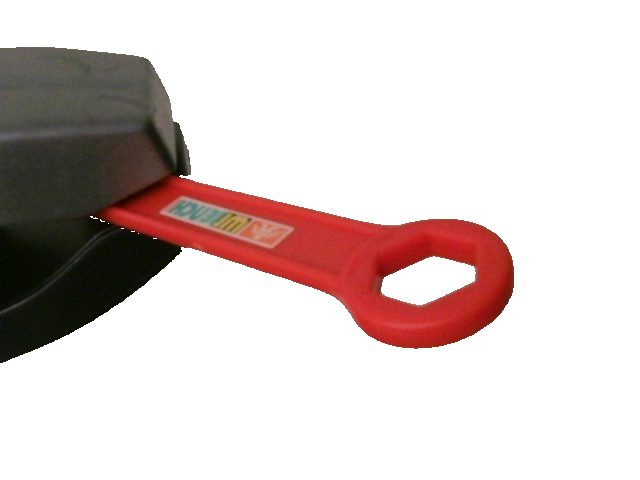}}%
\fbox{\includegraphics[width=0.241\linewidth]{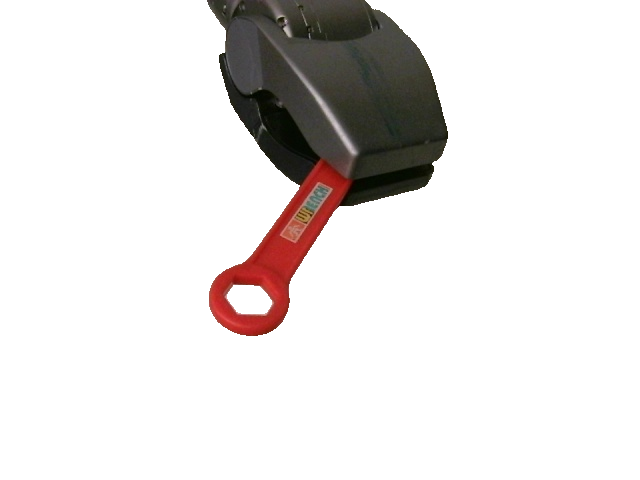}}%
\fbox{\includegraphics[width=0.241\linewidth]{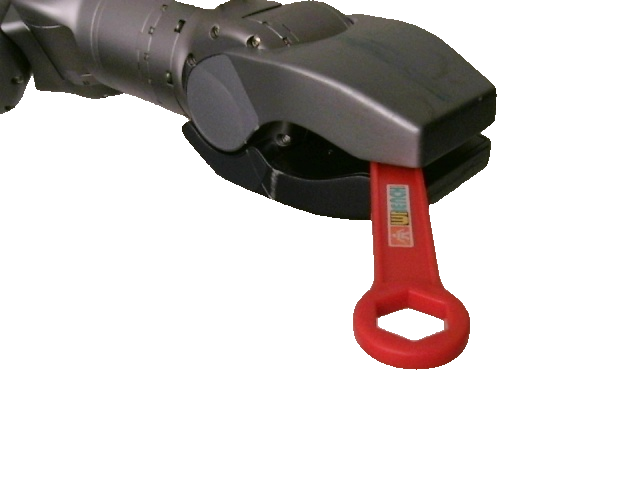}}%
\fbox{\includegraphics[width=0.241\linewidth]{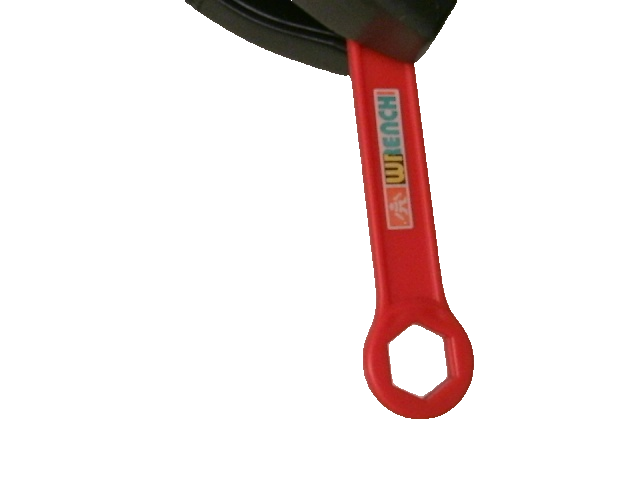}}%

\fbox{\includegraphics[width=0.49\linewidth]{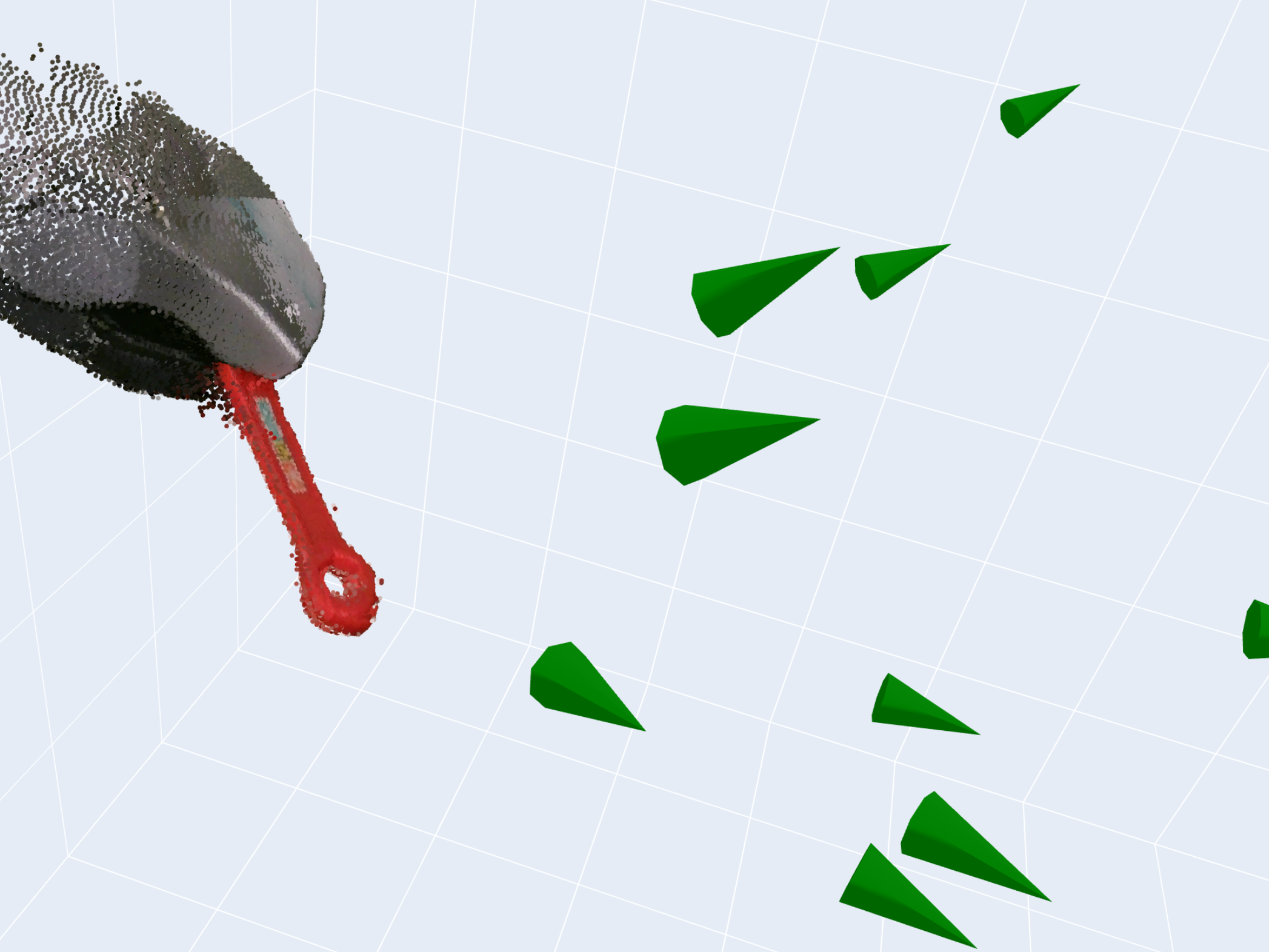}}%
\fbox{\includegraphics[width=0.49\linewidth]{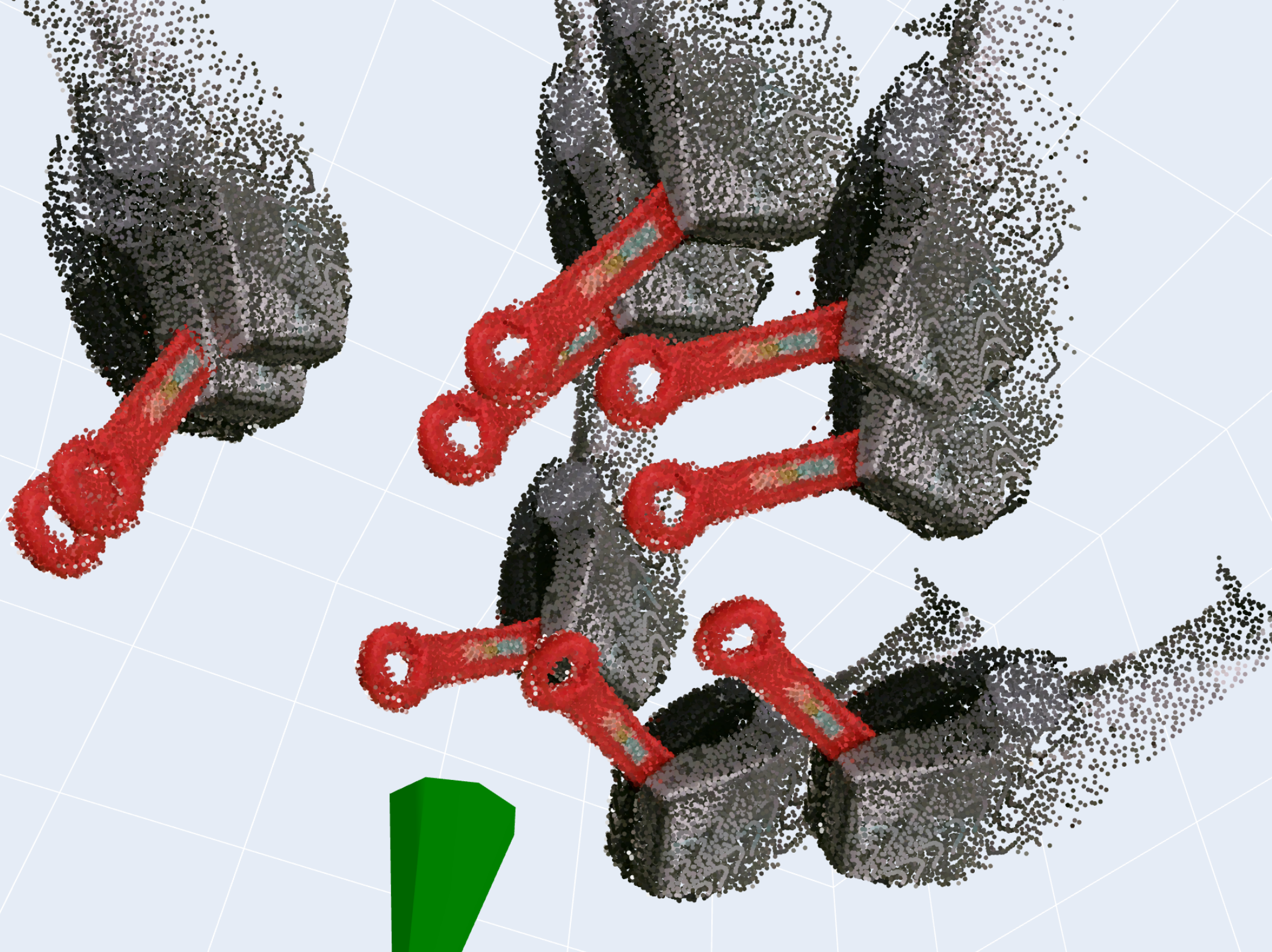}}%
\caption{(\underline{Top}) Examples of images of a robot gripper holding a toy wrench with masked white backgrounds. A total of nine such images are inputted into the 3D foundation model. (\underline{Bottom-left}) The structure from motion and multi-view stereo solution estimates a dense reconstruction and camera poses. This includes a reconstruction in a single pose and cameras at nine poses. (\underline{Bottom-right}) We can instead recover the object pose estimation and dense reconstruction solution by assuming a fixed camera. This includes a camera at a single pose with reconstructions at nine poses. All cameras are illustrated as green cones. The dense reconstructions have been down-sampled to enable efficient visualisation.}\label{fig:sfm_to_ope}
\end{figure}

\textbf{Transformation Notation:}
Coordinate transformations frequently appear throughout this paper. Here, a transformation matrix in the Special Euclidean Group,  $T\in\mathrm{SE}(3)$, acts on a batch of 3D coordinates $Y$. For brevity, we shall shorthand the transformation of $Y$ by $T$ as a matrix multiplication $YT^{\top}$. In practice, to implement the transform, we need to convert the 3D points to homogeneous coordinates by concatenating a column of ones and multiplying by $T^{\top}$. Then, the first 3 columns of the product are taken. That is, we use the shorthand:
\begin{align}
YT^{\top}:=\{[Y,\bb{1}]T^{\top}\}_{[:,1:3]},
\end{align}
where $\bb{1}$ denotes a column of ones, and $\{\cdot\}_{[:,1:3]}$ denotes taking the first three columns.
\section{Joint Geometry and Pose Estimation}\label{sec:GPE}
\subsection{Object to Camera Pose Estimation and Unscaled Reconstruction}
DUSt3R is designed to perform Structure from Motion and Multi-view Stereo (SFM+MVS), i.e. estimating camera poses while reconstructing a 3D structure from images of a static scene taken at different camera poses. Here, we use 3D foundation models trained for SFM+MVS also to solve \emph{Object to Camera Pose Estimation and Unscaled Reconstruction} (OPE+UR) --- the problem of estimating the pose of a moving rigid object that is masked out, in the frame of a stationary camera, while producing a dense but unscaled 3D reconstruction. 

We capture a set of images $\{I_{1},\ldots,I_{N}\}$, using a fixed external RGB camera, of the gripper and grasped object moving to several different poses. We also compute binary segmentation masks that identify the robot's gripper with the held object and remove other contents in the image. We then use the segmentation masks to zero out the confidence of masked-out pixels, then solve \cref{eqn:opt} and obtain dense unscaled reconstruction $\bar{X}_{Den}$ along with relative camera poses $\{\bar{P}_{1},\ldots,\bar{P}_{N}\}$. By enforcing the external camera to be stationary, we can then transform the reconstruction at the $n^{th}$ pose, denoted as $[\bar{X}_{Den}]^{n}$, by applying the transform $\bar{P}_{n}^{-1}$ to each point in the reconstruction, i.e.,
\begin{equation}
[\bar{X}_{Den}]^{n}:=\bar{X}_{Den}(\bar{P}_{n}^{-1})^{\top}. \label{eqn:proj_dense}
\end{equation}
An example of inputting masked images of a gripper with a held wrench, moved to several poses is given in \cref{fig:sfm_to_ope}. The SFM+MVS solution from the foundation model is given in the bottom-left subfigure, while the equivalent OPE+UR solution is shown in the bottom-right. The SFM+MVS solution assumes that the scene, in this case, the gripper and wrench, is static while the camera moves. Alternatively, the OPE+UR solution assumes that the camera is stationary and the gripper and wrench move.

Two remaining challenges prevent the direct use of the reconstruction and object poses from being used in robotics applications:
\begin{enumerate}
\item both the reconstruction and pose translations are of relative, and not to metric scale. 
\item the poses are not in the frame of the robot. Additionally, we assume that the pose of the external camera with respect to the robot has not been carefully calibrated, and so is not known in advance.
\end{enumerate}
Next, we shall elaborate on how to recover both metric scale and transformations into the robot's coordinate frame.

\begin{figure}[t]
    \centering
\includegraphics[width=0.8\linewidth]{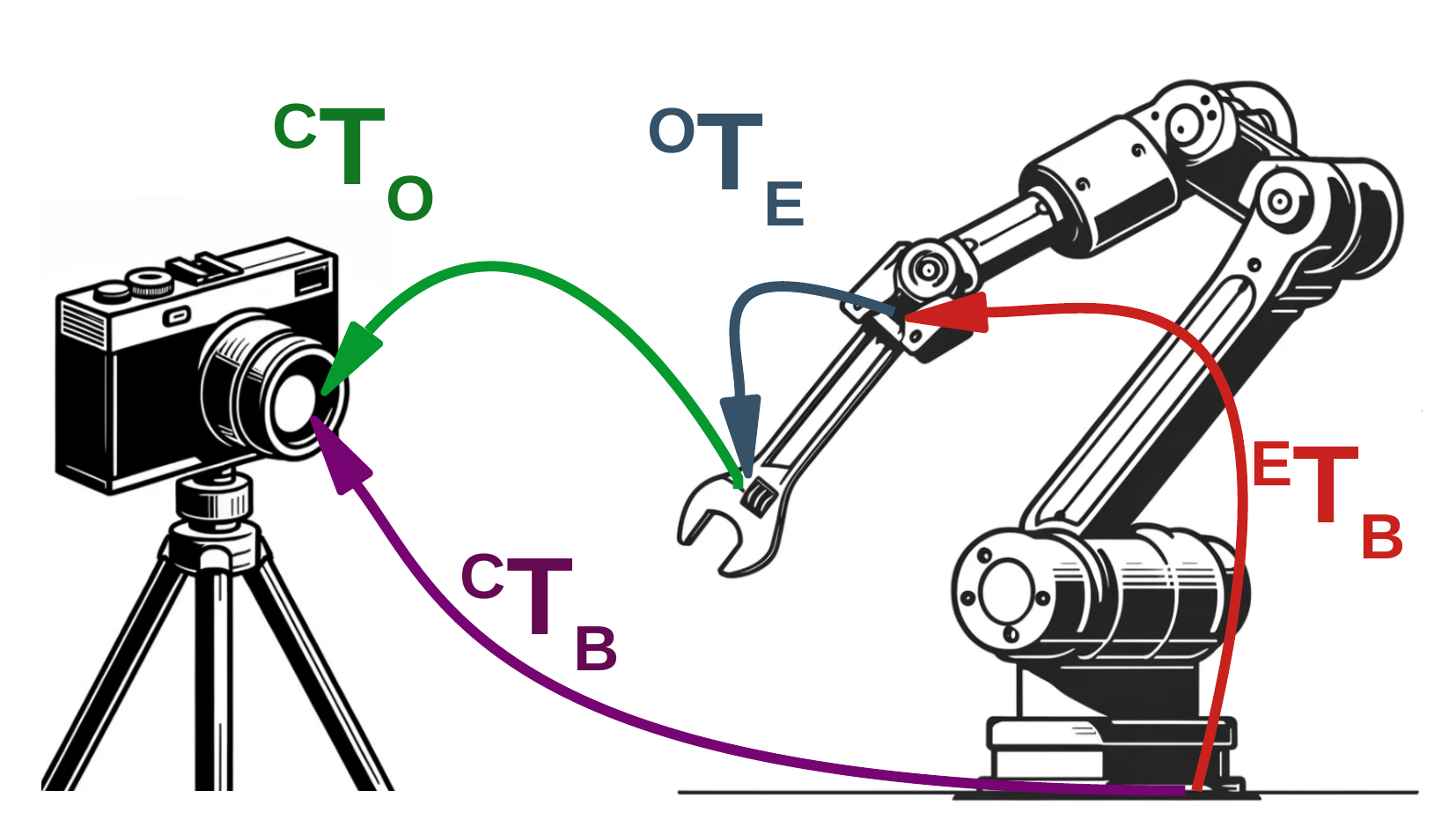}
    \caption{We define the transformations $^{E}T_{B}$, $^{O}T_{E}$, $^{C}T_{O}$, $^{C}T_{B}$ between robot's base, end-effector, the object and the camera.}
    \label{fig:transformations}
\end{figure}
\begin{figure}[t]
\centering
\begin{tikzpicture}
  \node (left) at (0,0) [rounded corners, draw] {$[^{C}T_{O}]_{n}$};
  \node (right1) at (5,1.25) [rounded corners, draw] {$[^{C}T_{O}]_{1}$};
  \node (right2) at (5,0.5) [rounded corners, draw] {$[^{C}T_{O}]_{2}$};
  \node (ellipsis) at (5, -0.225) {$\bb{\vdots}$};
  \node (right3) at (5,-1.) [rounded corners, draw] {$[^{C}T_{O}]_{N}$};
  
  \draw (left) -- (right1) node[midway, above] {$H^{-1}A_{n,1}H$};
  \draw (left) -- (right2) node[midway, below] {$H^{-1}A_{n,2}H$};
  \draw (left) -- (right3) node[midway, below] {$H^{-1}A_{n,N}H$};
\end{tikzpicture}
\caption{Each ${[^{C}T_{O}]}_{n}$ is connected to the other $^{C}T_{O}$ transforms by pre-multiplying $(H^{-1}A_{n,m}H)$ to ${[^{C}T_{O}]}_{m}$, where $m$ denotes the index of the other $^{C}T_{O}$ matrices. We can then represent $[^{C}T_{O}]_{n}$ by averaging the incoming results from all other ${[^{C}T_{O}]}$.}\label{fig:connected_graph}
\end{figure}
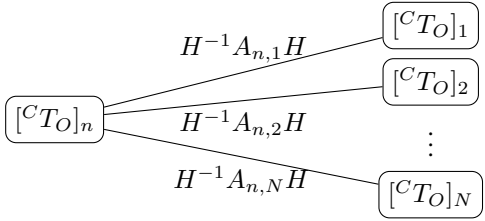
\subsection{Coordinate-alignment Problem}
As the robot's gripper is being moved to a different pose while images are being captured, we can also record all $N$ end-effector poses $\{E_{1},\ldots, E_{N}\}$. By considering the robot's base $B$, the end-effector, the object and the camera, we can define the rigid transformations $^{E}T_{B}$, $^{O}T_{E}$, $^{C}T_{O}$, $^{C}T_{B}$. These are illustrated in \cref{fig:transformations}. We note that $^{E}T_{B}$ and $^{C}T_{O}$ change as the gripper and object are moved around to the $N$ different poses. Out of the defined transforms: 
\begin{itemize}
    \item $^{E}T_{B}$ transforms are known and given by the recorded end-effector poses. We denote the transformation at the $n^{th}$ pose as,
    \begin{align}
    [^{E}T_{B}]_{n}=E_{n}, && \text{for} \quad n=1,\ldots,N.
    \end{align}
    \item $^{C}T_{O}$ transforms have rotations given by rotations of the estimated object poses, but the translation is proportional to the estimated translation up to an unknown scale $\alpha$. We define the transformation at the $n^{th}$ pose as:
    \begin{align}
    &[^{C}T_{O}]_{n}(\alpha)=\begin{bmatrix}R_{n} & \alpha\bb{t}_{n}\\
    \bb{0} & 1
    \end{bmatrix},\\
    &\text{where }\bar{P}_{n}^{-1}=\begin{bmatrix}R_{n} & \bb{t}_{n}\\
    \bb{0} & 1
    \end{bmatrix},  \text{ for } n=1,\ldots,N.
    \end{align}
    Here, we denote $\bb{0}$ as a row vector with 3 zeros.
    \item $^{O}T_{E}$ is the unknown rigid transformation we seek to solve. This remains fixed as the end-effector moves around in its workspace.
    \item $^{C}T_{B}$ is also assumed to be unknown as we do not require calibrating camera extrinsics. This can be evaluated by considering the transformations from the base to the end-effector, then to the object, and finally to the camera. That is, 
    \begin{align}
    ^{C}T_{B}={^{E}T_{B}} {^{O}T_{E}} ^{C}{T}_{O}.
    \end{align}
\end{itemize}
We shall solve to find $\alpha$ and $^{O}T_{E}$, leveraging the fact \textbf{the camera remains stationary relative to the robot's base as the gripper moves to different poses}. That is, $^{C}T_{B}$ does not change between end-effector movements. Therefore, we have the equalities:
\[
\left.\begin{array}{rl}
^{C}T_{B} = & {[^{E}T_{B}]_{1}} {[^{O}T_{E}]} {[^{C}T_{O}]_{1}(\alpha)} \\
& {[^{E}T_{B}]_{2}} {[^{O}T_{E}]} {[^{C}T_{O}]_{2}(\alpha)} \\
           & \quad \vdots \\
           = & {[^{E}T_{B}]_{N}} {[^{O}T_{E}]} {[^{C}T_{O}]_{N}(\alpha)}
\end{array}\right\} \text{N equalities}
\]

\begin{figure}[t]
\centering
\fbox{\includegraphics[width=0.48\linewidth]{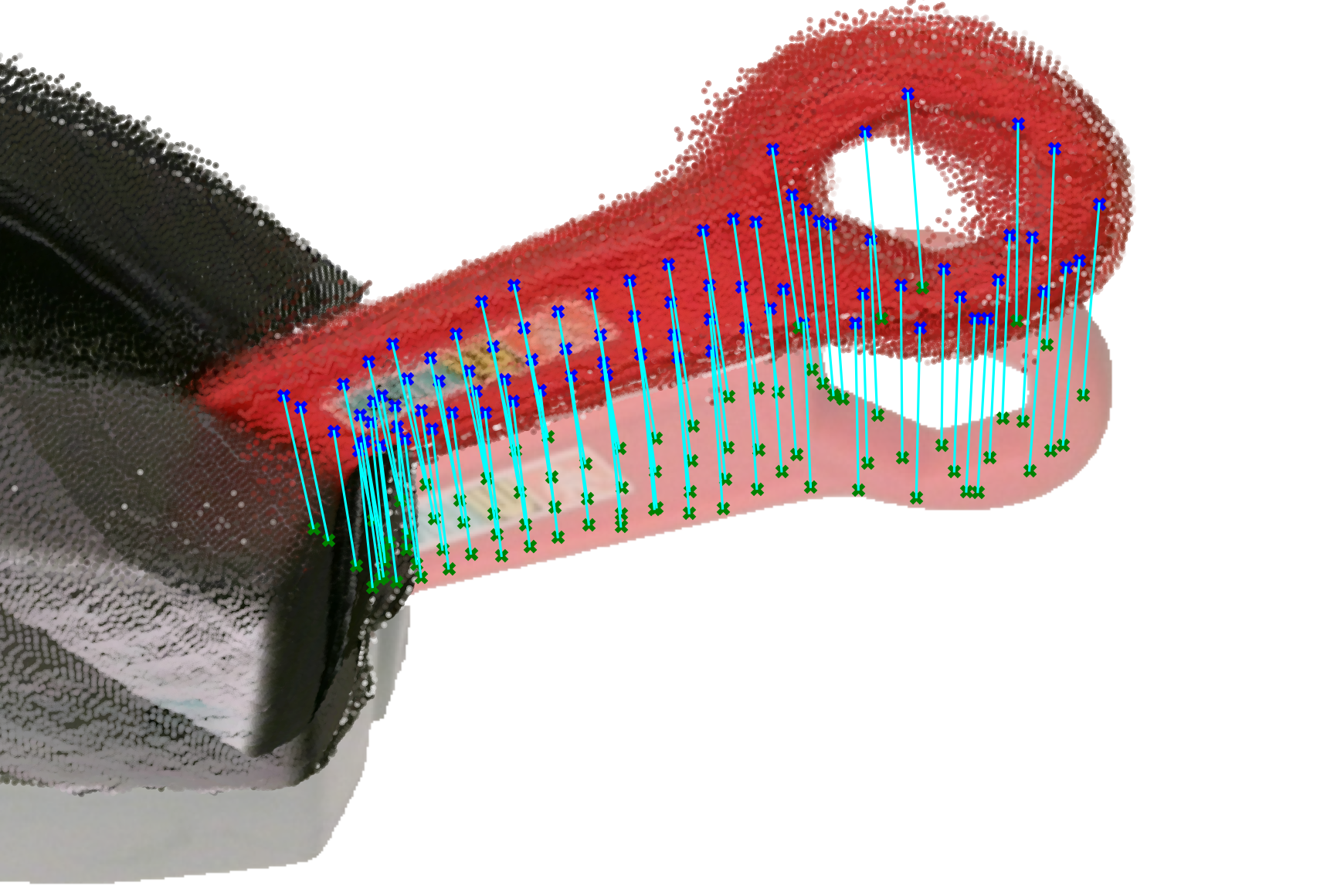}}%
\fbox{\includegraphics[width=0.48\linewidth]{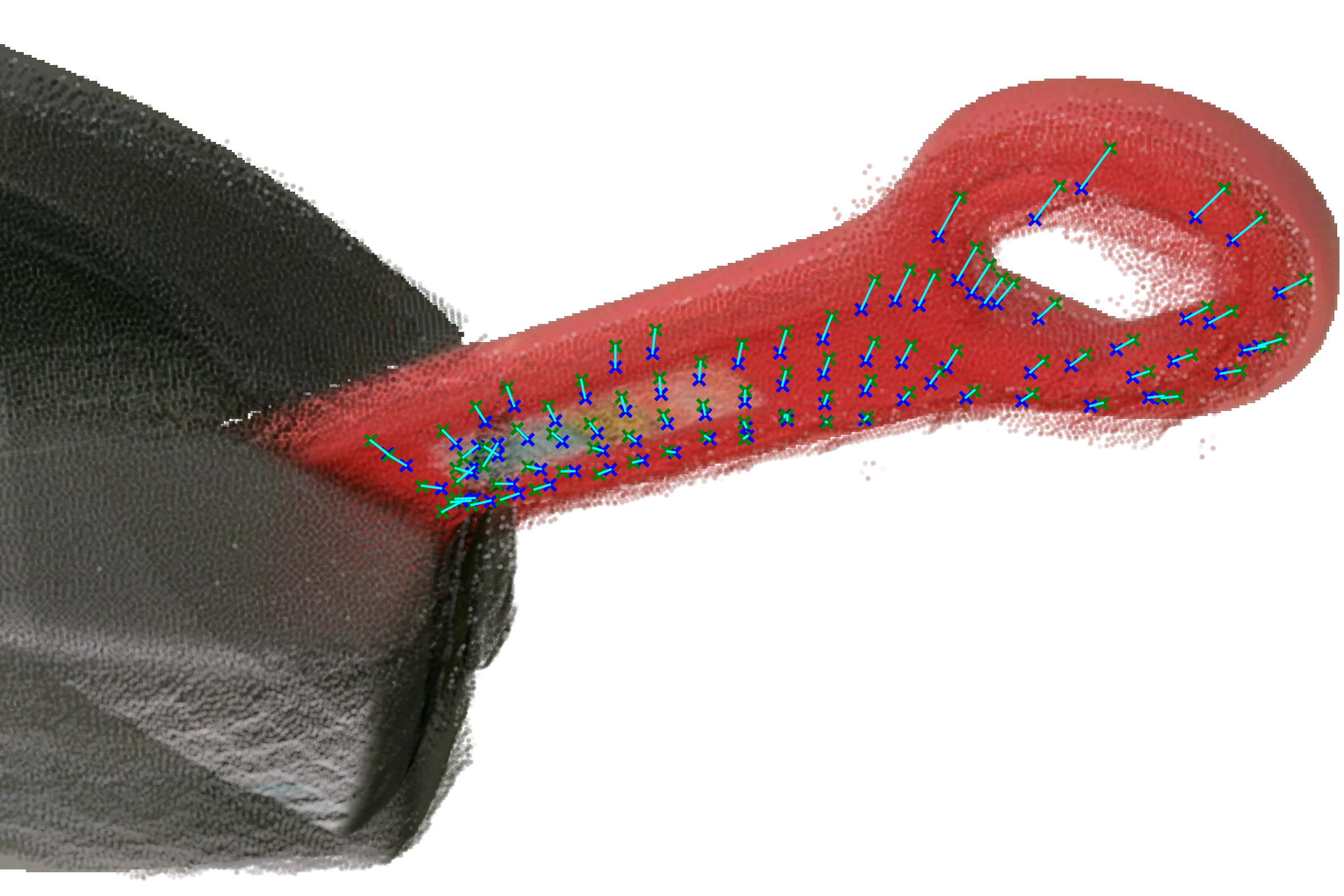}}
\caption{We minimise the distances between the dense reconstructions, transformed under the estimated $f_{i}$ and $\hat{P}_{i}^{-1}$ rendered into the camera's view in 2D. (\underline{Left}) We show an example of the dense projected by $f_{i}$ illustrated with a small selected set of sampled points in blue. We have the points under the $\hat{P}_{i}^{-1}$ transformation, corresponding to the blue samples shown in green, overlaid on a ground-truth image. The distances between the correspondences are shown as cyan lines. (\underline{Right}) After optimising until convergence, distances between the illustrated selected points are minimal. The estimation projected into the camera's view almost entirely aligns and overlaps with the ground-truth image.}\label{fig:alignment}
\end{figure}

For a given $[^{C}T_{O}]_{n}$, $n=1,\ldots,N$, we can thus use these equalities to represent it by a combination of other $[^{C}T_{O}]_{m}$, where $m=1,\ldots,N$ and $m\neq n$. This is illustrated in \cref{fig:connected_graph}. For greater clarity in notation, let us define the notation: 
\begin{align}
H:=^{O}T_{E}, && A_{n,m}:={[^{E}T_{B}]_{n}}^{-1}{[^{E}T_{B}]_{m}}. 
\end{align}
We then construct the parameterised model $f_{n}(H,\alpha)$ to represent $[^{C}T_{O}]_{n}(\alpha)$ as,
\begin{align}
f_{n}(H,\alpha):=\frac{1}{N-1}\sum_{m}(H^{-1}A_{n,m}H) [^{C}T_{O}]_{m}(\alpha).\label{eqn:estimator}
\end{align}
We seek to minimise the $l2$ distance between the dense points after transformation by $f_{n}(H,\alpha)$ and the point-set $[\bar{X}_{Den}]^{n}$, \textbf{rendered to the camera}. Specifically, $[\bar{X}_{Den}]^{n}$, introduced in \cref{eqn:proj_dense}, refers to the dense reconstruction under the transformation $\bar{P}_{n}^{-1}$. Here, we note that the physical scale factor $\alpha$ does not impact how the dense points are rendered in the camera frames. However, as the end-effector transformations contain absolute translations and are incorporated into the definition of $f_{n}$, we can recover $\alpha$ via the optimisation.

To render the reconstruction, let $K\in\mathbb{R}^{3\times 3}$ be the intrinsic matrix of the camera, then we define the projection $F$ as:
\begin{align}
F(K,\bar{X})\!:=\!\{(\frac{u}{w}\!,\!\frac{u}{w})\lvert \begin{bmatrix}u\\v\\w\end{bmatrix}\!=\!K\bb{x}^{\top}, \bb{x} \text{ is a row of }\bar{X}\!\}, 
\end{align}
which returns a set of 2D-pixel coordinates. We define the loss as the mean absolute error (MAE) between the dense points, under the $f_{n}(H,\alpha)$ and $\bar{P}_{n}^{-1}$ transformations, projected back to the camera's view, and iterating through each $n$. That is: 
\begin{align}
Y^{n}(H,\alpha):=\hat{X}_{Den}[f_{n}(H,\alpha)]^{\top},\\
\ell(H,\alpha)\!=\!\mathrm{MAE}\{F(K,[\bar{X}_{Den}]^{n}),F(K,Y^{n}(H,\alpha))\}.
\end{align}
Here, $Y^{n}$ is the set of dense points transformed by the estimated $f_{n}$. The coordinate alignment problem is then:
\begin{align}
\arg\min_{H,\alpha}\ell(H,\alpha)\\
\text{S.T.} \quad H\in\mathrm{SE}(3),\\
\alpha>0.
\end{align}
We apply first-order optimisers and ensure $H$ and $f_{n}$ are valid transformations in $\mathrm{SE}(3)$ by projecting the rotation components onto the special orthogonal group $\mathrm{SO}(3)$ via the Procrustes implemented in the RoMa library \cite{bregier2021deepregression}.

An illustration of the optimisation with camera-view distances is given in \cref{fig:alignment}. The dense reconstruction transformed by $f_{i}$ is rendered. We select a small set of sample points on rendered reconstruction (displayed in blue) and find the corresponding points on $[\bar{X}_{Den}]^{n}$ (displayed in green), and indicate their distances in cyan. The left subfigure shows a snapshot during optimisation, while the right subfigure shows a snapshot when optimisation has converged. We observe that the distances between corresponding points are minimal, and the rendered dense reconstruction overlaps tightly with the underlying ground-truth image.



\subsection{From Robot Configurations to Object Coordinates}
After solving the coordinate-alignment problem described in the previous subsection, we can scale our dense reconstruction by the obtained factor $\alpha$. We seek to derive the mapping, which we denote as $\psi$, that takes as input the robot's $d$-dimensional configuration $\bb{q}\in\mathcal{Q}\subseteq\mathbb{R}^{d}$, where $\mathcal{Q}$ denotes the robot's \emph{configuration space}, and outputs the 3D positions of points of interest in the frame of the robot. Typically, the configuration space of a robot manipulator is its joint space. We represent the points of interest, on which we wish to track in the robot's frame, as $X'\in\mathbb{R}^{N_{POI}\times 3}$ where we allow $N_{POI}$ points. 

Let us define the forward kinematics of the robot as $\phi(\bb{q})\in\mathrm{SE}(3)$. Then, the coordinates of the points of interest in the frame of the robot's base, $[X']_{B}$, by simply evaluating the inverse sequence of transformations,
\begin{align}
[X']_{B}=\psi(\bb{q})=X'((\phi(\bb{q}) H)^{-1})^{\top}.
\end{align}
We can also define the inverse, $\psi^{-1}$, which takes $[X']_{B}$ as input and finds corresponding configuration. Let us define the inverse kinematics of the robot as $\phi^{-1}(\bb{x})\in\mathcal{Q}$, then,
\begin{align}
\psi^{-1}([X']_{B})=\phi^{-1}([X']_{B}(H)^{\top}). \label{eqn:inv_mapping}
\end{align}
Therefore, we have both forward and inverse mappings from the robot's configuration to coordinates of specified points on the held object. This is achieved by leveraging the forward and backward mappings from the robot's configuration to its end-effector, and then to points on the held object.


\section{Empirical Evaluations}\label{sec:empirical}
We empirically investigate the effectiveness of our joint geometry and pose estimation framework. We are particularly interested in answering the following questions: (1) How accurate are solutions obtained in GPE? This is addressed in \cref{subsec:gen_results}. (2) Can GPE work robustly when the number of collected RGB images is severely limited? This is addressed in \cref{subsec:limited_data}. (3) Can we use the estimated object geometry and pose and shape the robot's motion relative to specified coordinates on the object? This is addressed in \cref{subsec:shaping}.

\subsection{Experimental Setup, Metrics and Baselines}
\textbf{Setup:} We evaluate our framework on six different objects, including a (1) hammer; (2) screwdriver; (3) wrench; (4) wooden block; (5) roll of tape; (6) brush. For each object, we collect a set of 9 images of the robot manipulator with the object in-hand, at different configurations in front of the camera. We use the Unitree Z1 manipulator, and a low-cost (below $15$ USD) USB camera. The image backgrounds are then removed via language-guided image segmentation, using Segment Anything \cite{kirillov2023segany}. The resulting images are inputted into our joint object geometry and pose framework, and outputs for each object are evaluated against held-out test sets with 4 images each. Note using fewer than 10 images is generally considered a challenging \emph{sparse} 3D vision task. Established methods such as COLMAP, which are critical in initialising NeRF \cite{mildenhall2020nerf} and Gaussian Splatting models \cite{splatting_GS}, would not provide meaningful solutions for image sets of this size.

\begin{figure}[t]
\centering
\fbox{\includegraphics[width=0.32\linewidth]{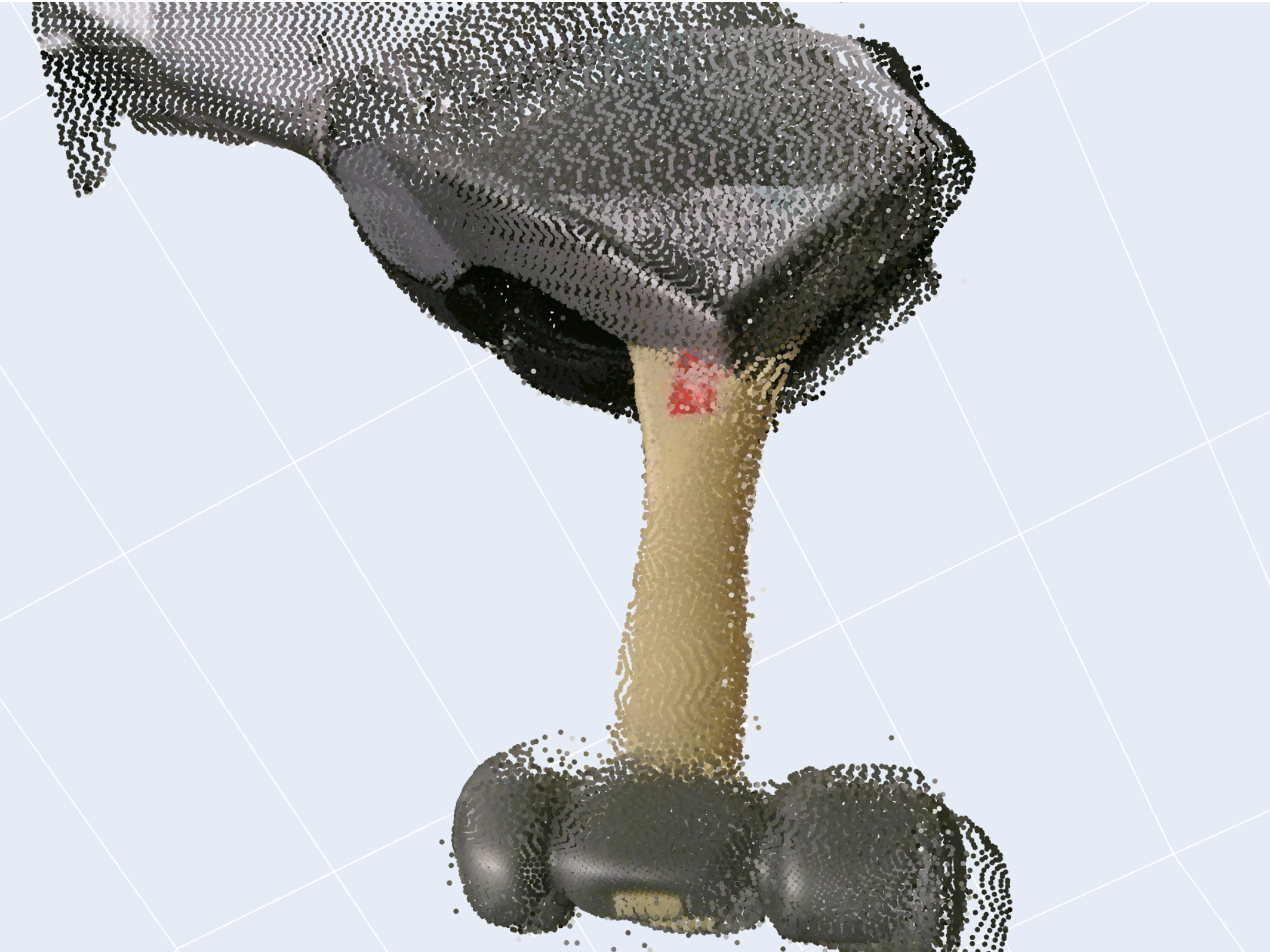}}%
\fbox{\includegraphics[width=0.32\linewidth]{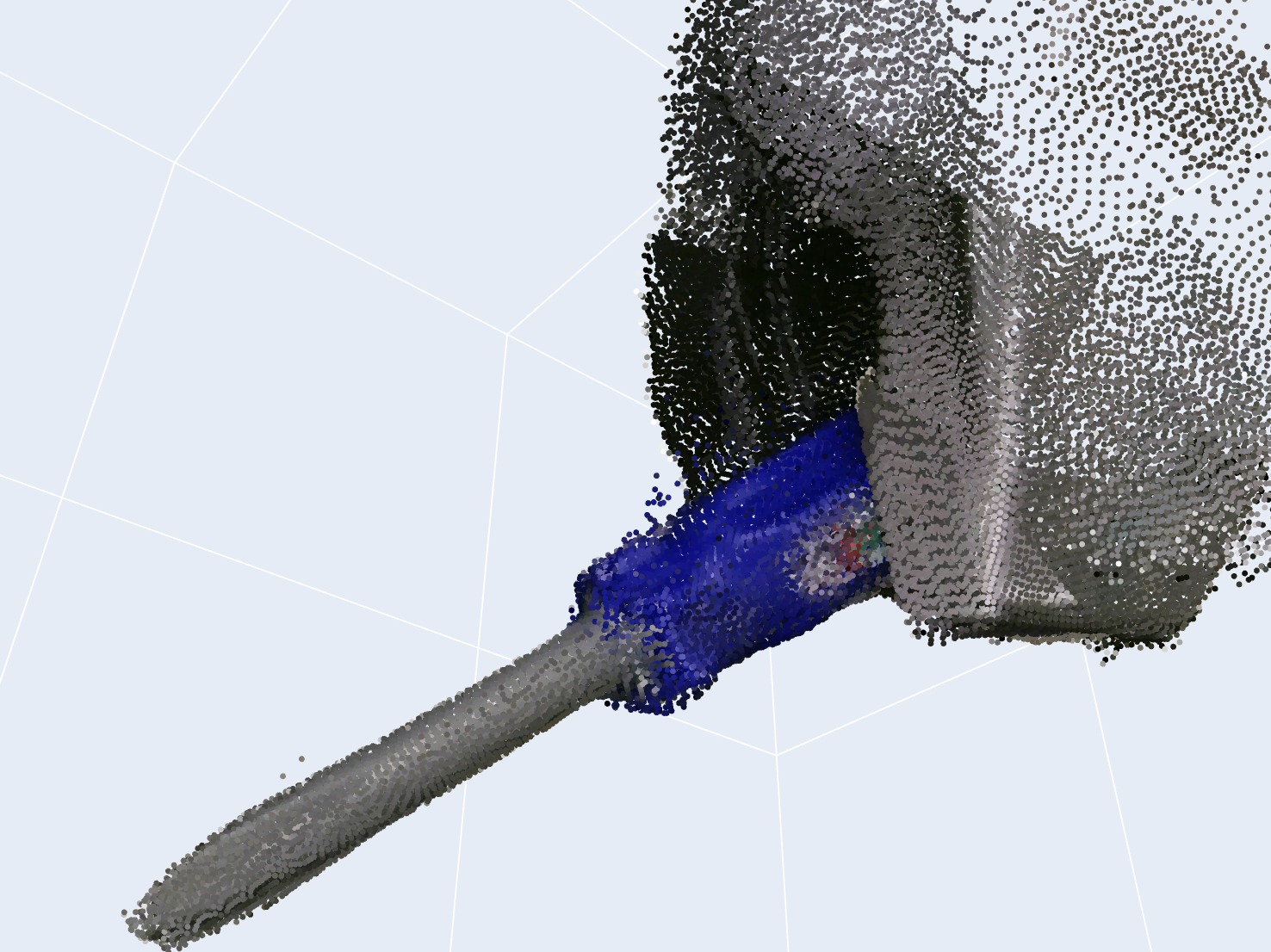}}%
\fbox{\includegraphics[width=0.32\linewidth]{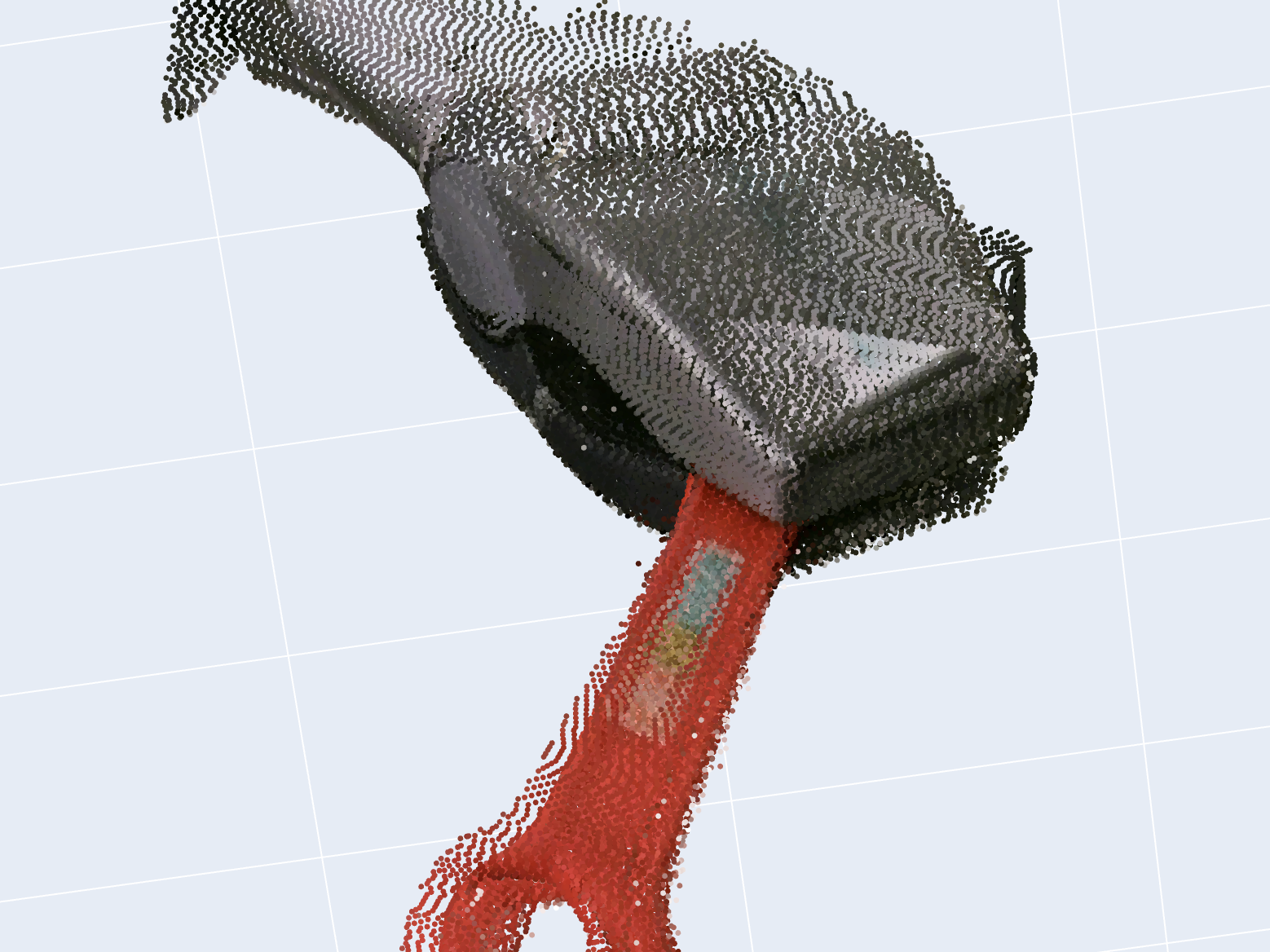}}%

\fbox{\includegraphics[width=0.32\linewidth]{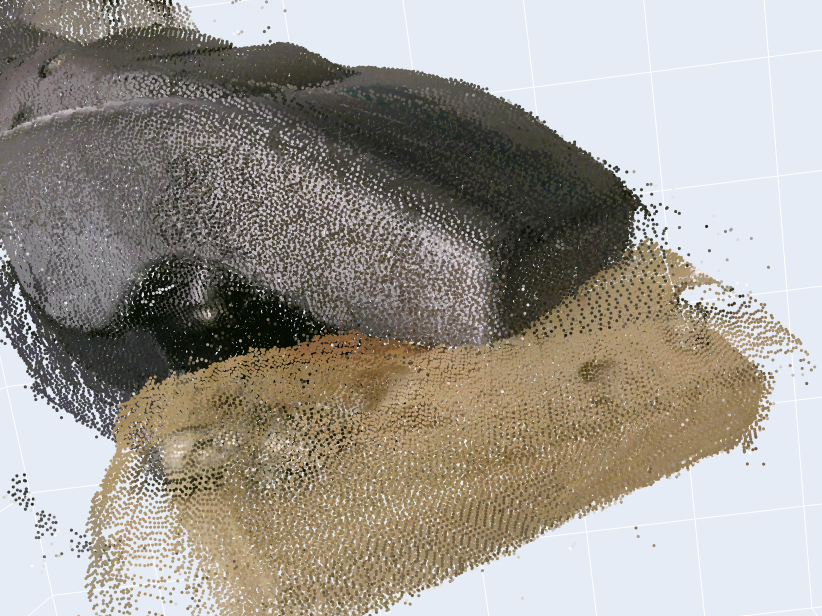}}%
\fbox{\includegraphics[width=0.32\linewidth]{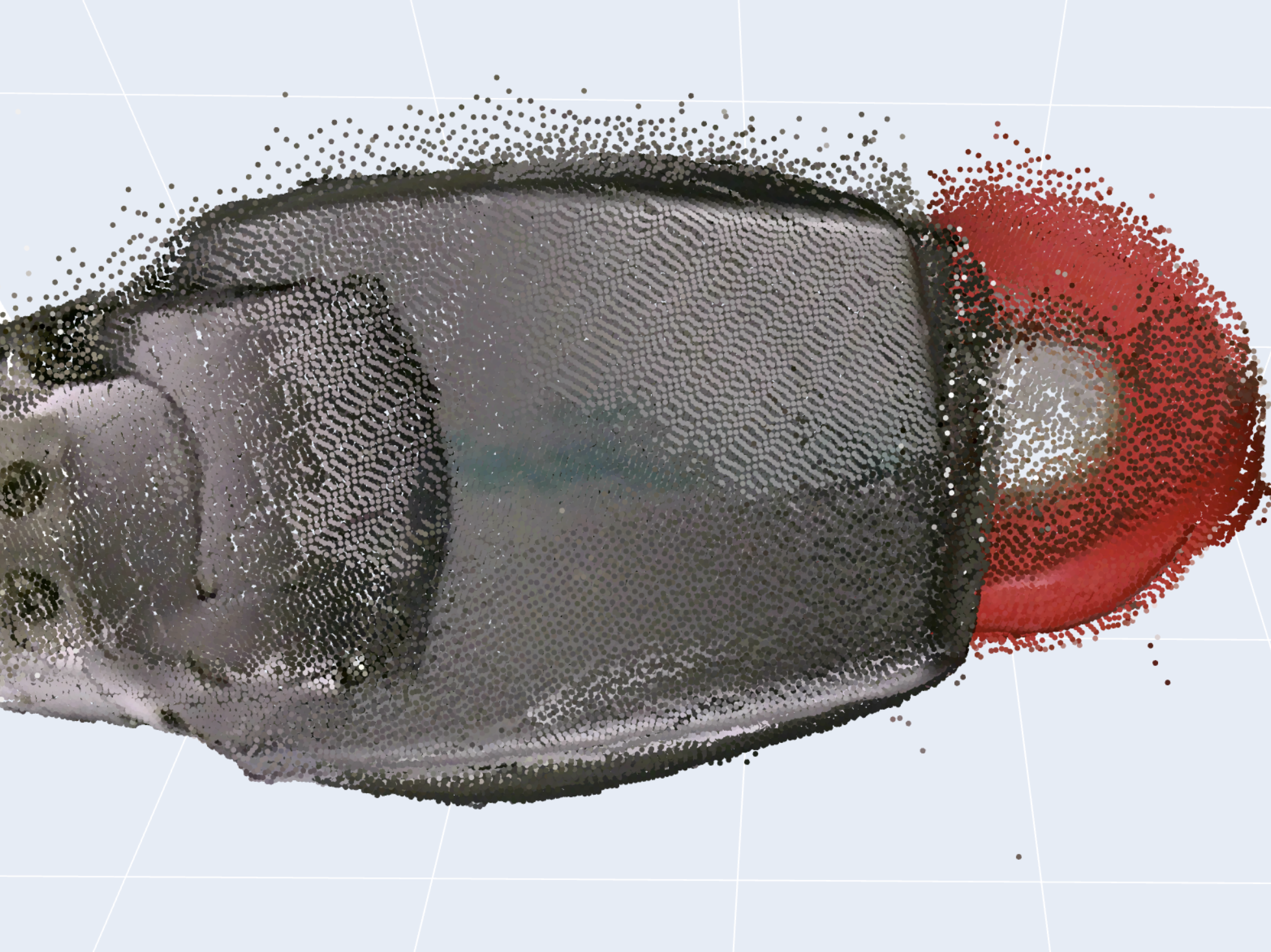}}%
\fbox{\includegraphics[width=0.32\linewidth]{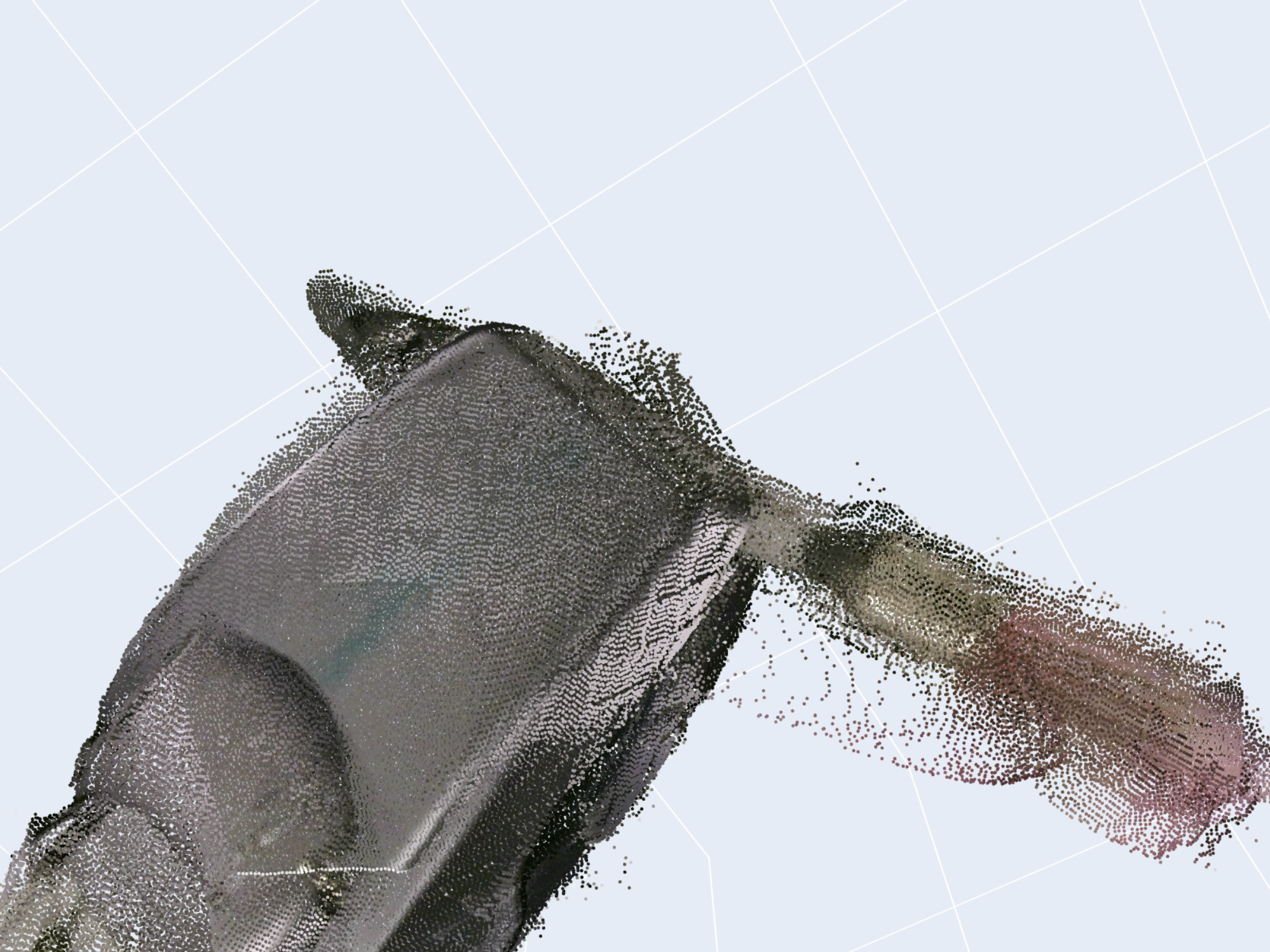}}%
\caption{Visualisations of the reconstructions of the evaluated objects.}\label{fig:recon}
\end{figure}

\begin{figure*}[t]
\centering
\includegraphics[width=0.32\linewidth]{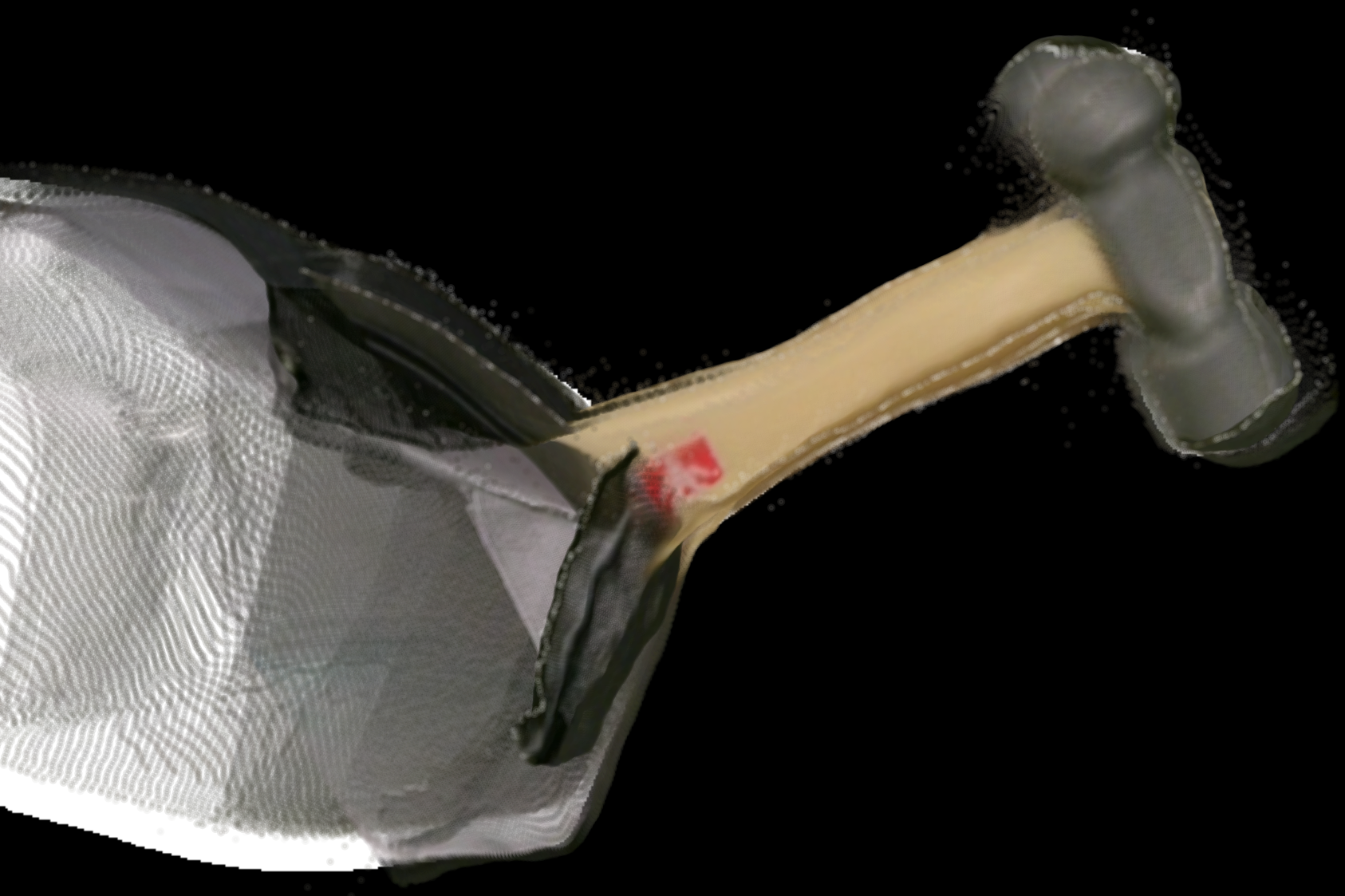}%
\hspace{0.02em}
\includegraphics[width=0.32\linewidth]{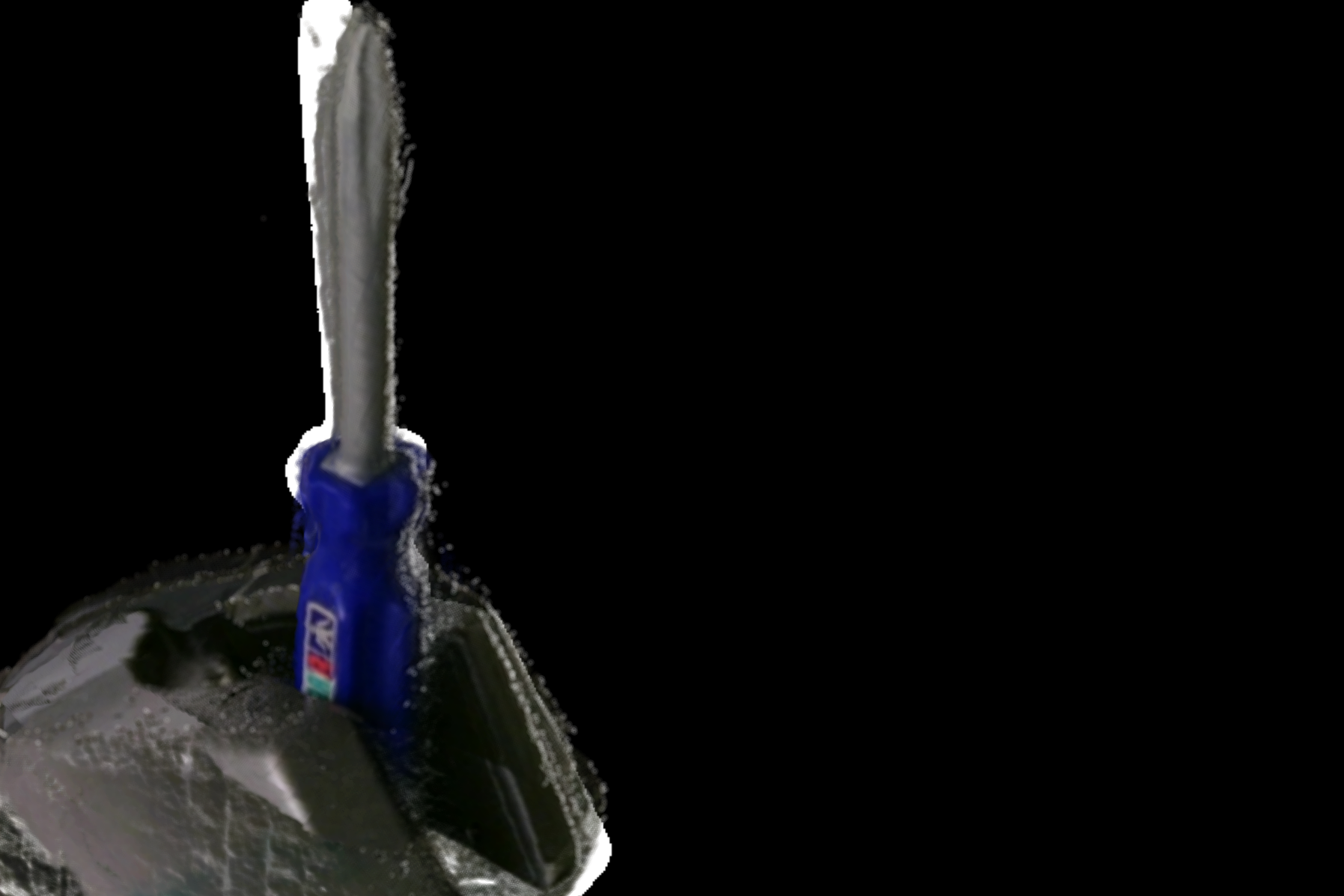}%
\hspace{0.02em}
\includegraphics[width=0.32\linewidth]{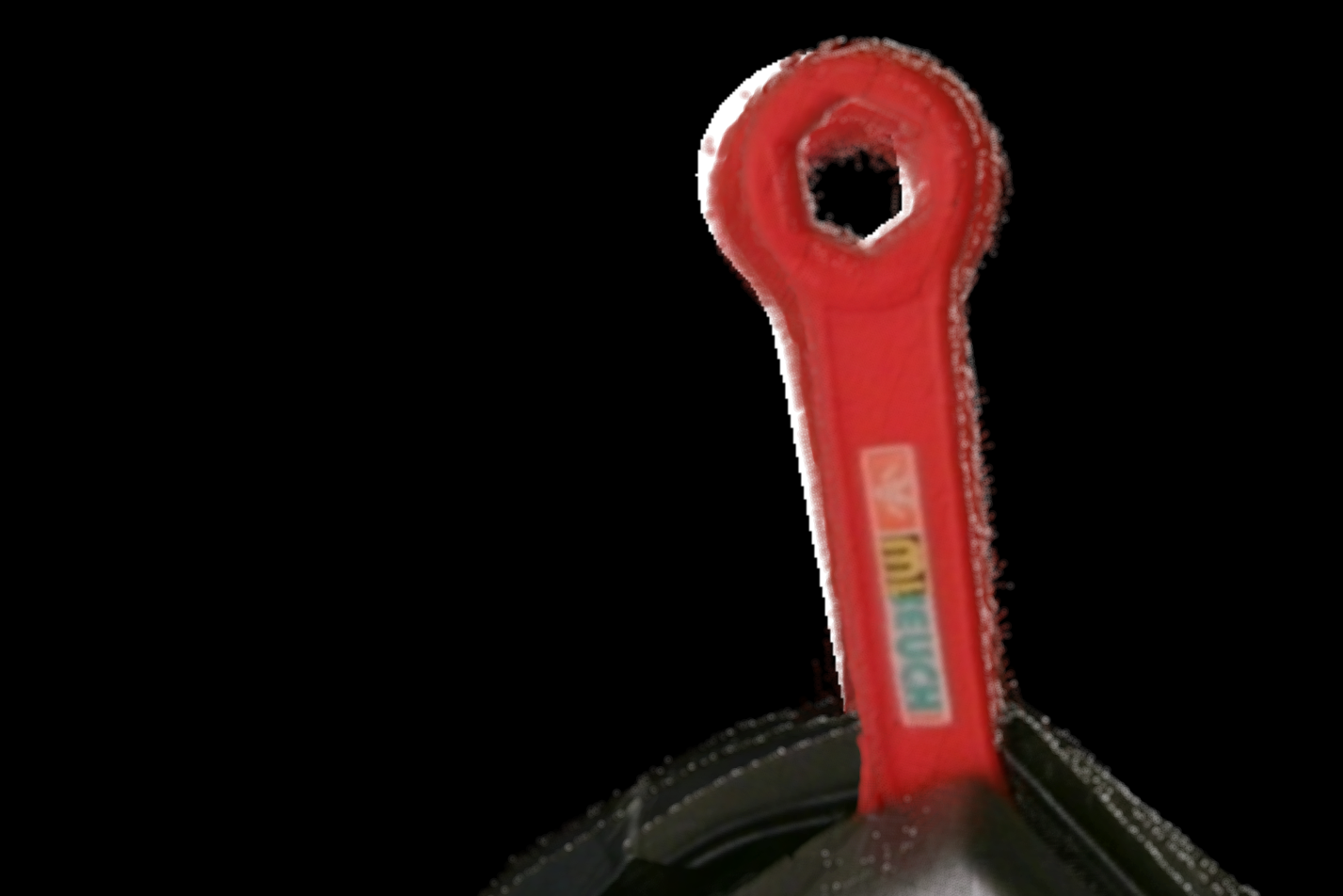}%

\vspace{0.2em}
\includegraphics[width=0.32\linewidth]{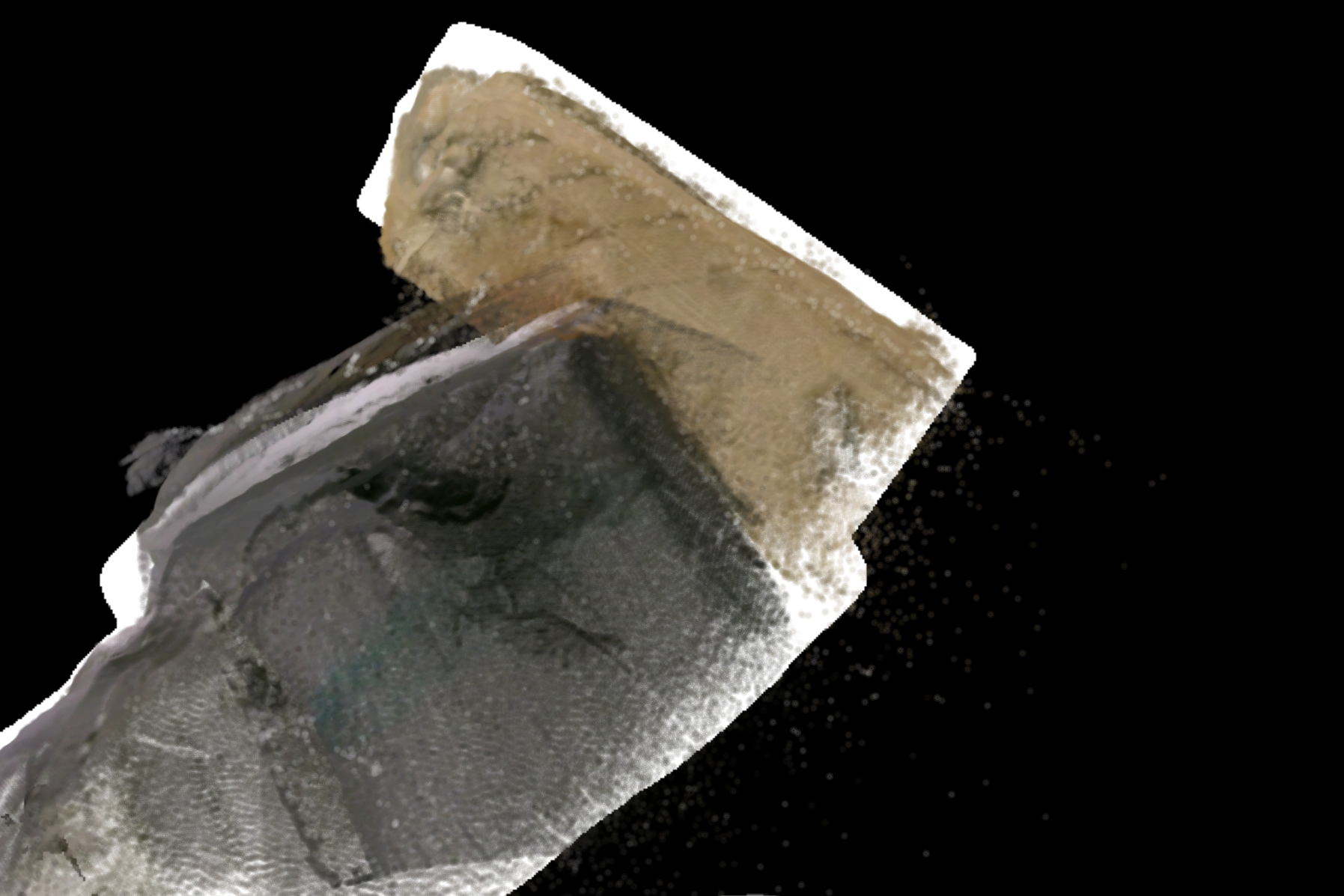}%
\hspace{0.02em}
\includegraphics[width=0.32\linewidth]{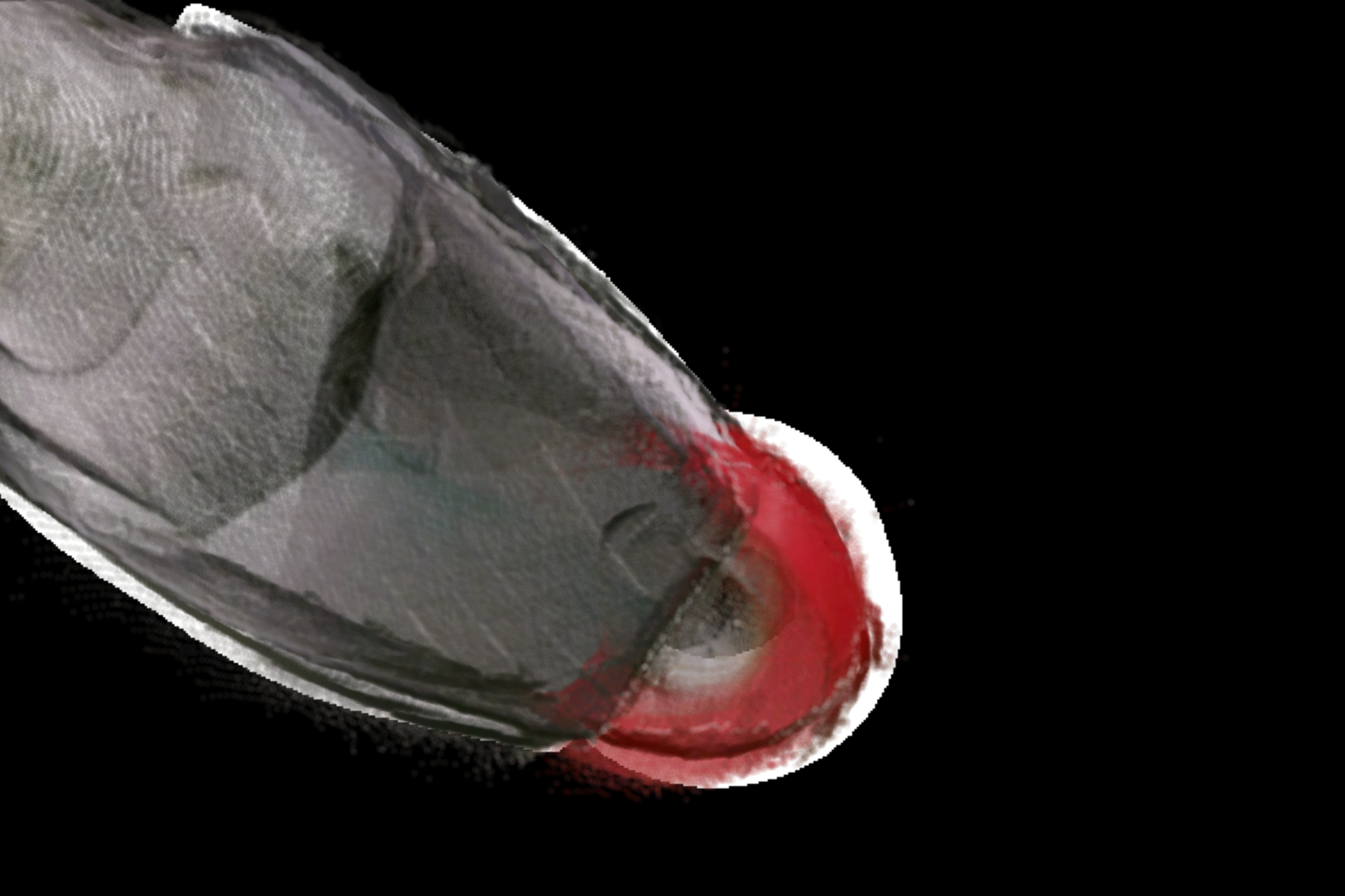}%
\hspace{0.02em}
\includegraphics[width=0.32\linewidth]{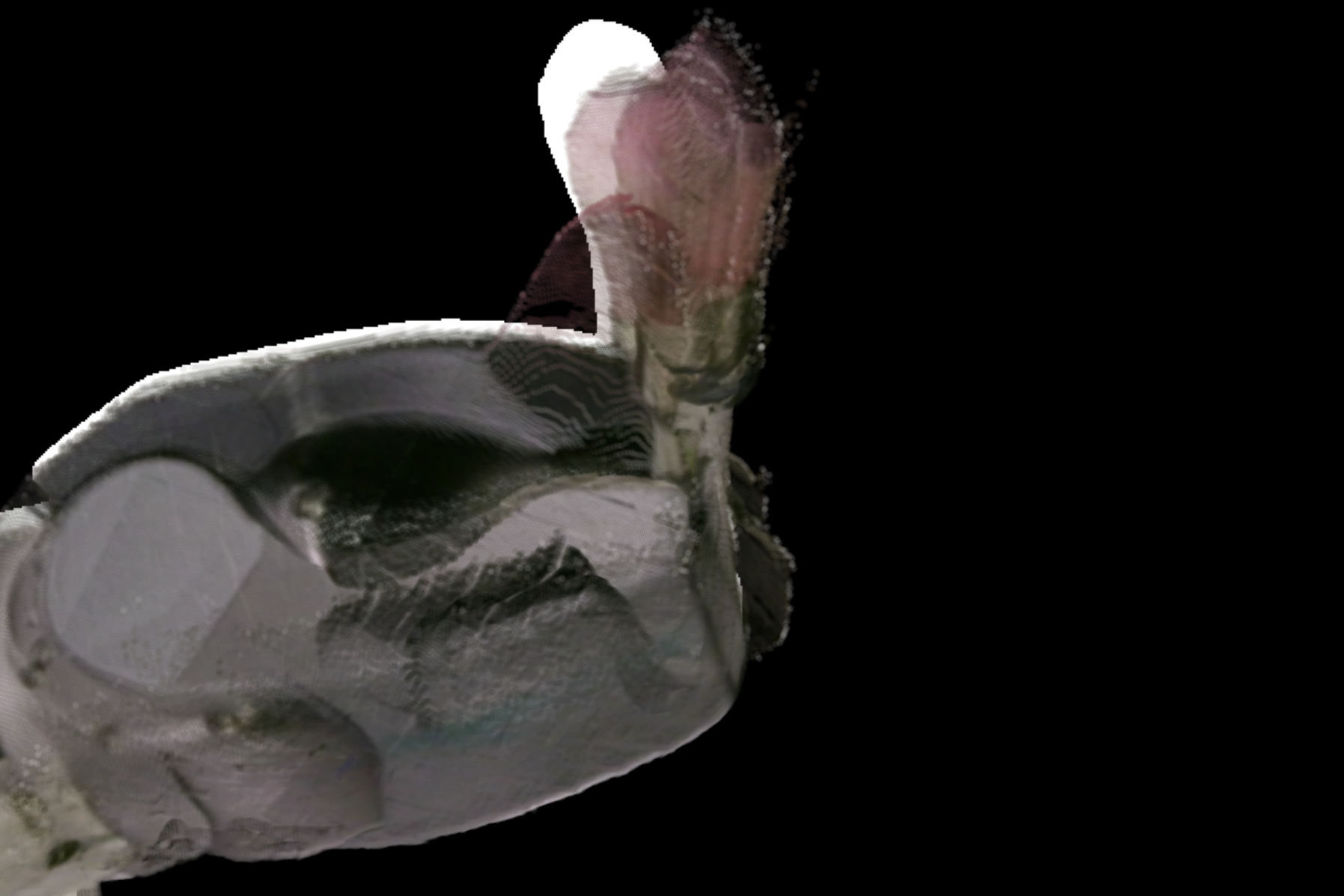}%

\caption{We overlay the dense reconstruction's camera projection on test configurations, with the test set ground truth silhouettes in white. We see that the reconstruction aligns with the silhouettes, highlighting the accuracy of our pose predictions.}\label{fig:results}
\end{figure*}

\begin{table*}[t]
\centering
\begin{adjustbox}{width=0.65\linewidth}
\begin{tabular}{@{}l|rrrrrr@{}}
\toprule[1pt]\midrule[0.3pt]
                         & \textbf{Hammer} & \textbf{Screwdriver} & \textbf{Wrench} & \textbf{Block} & \textbf{Tape} & \textbf{Brush} \\ 
                         \midrule
Structured (Ours)        & \textbf{1.42} & \textbf{1.16} & \textbf{1.11} & \textbf{0.69}         & \textbf{0.94}              & \textbf{0.52}           \\
Structured without Rendering & 2.03                       & 21.60                           & 8.11                       & 8.28                      & 16.85                    & 6.03                      \\
Unstructured Direct Regression        & 122.94                     & 92.53                           & 145.33                     & 27.05                     & 111.50                   & 85.42 \\
\midrule[0.3pt]\bottomrule[1pt]
\end{tabular}
\end{adjustbox}
\caption{Comparison between our approach and baselines in estimating the transformation between end-effector and in-hand objects. We report pixel distances (lowest in bold), for evaluations at test end-effector coordinates, between the reconstruction rendered to the camera view and the ground truth silhouettes.}
\label{tab:tool_handling}
\end{table*}

\begin{figure}[b]
\centering
\includegraphics[width=0.33\linewidth]{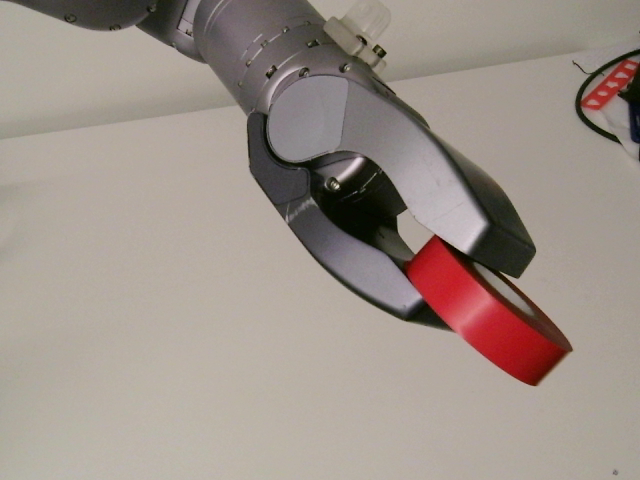}%
\includegraphics[width=0.33\linewidth]{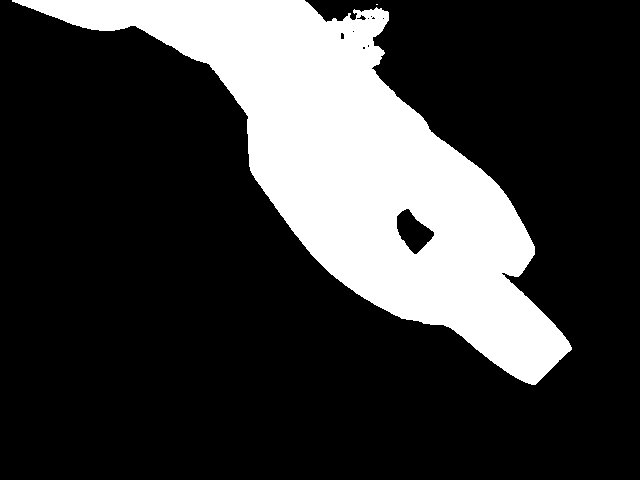}%
\includegraphics[width=0.33\linewidth]{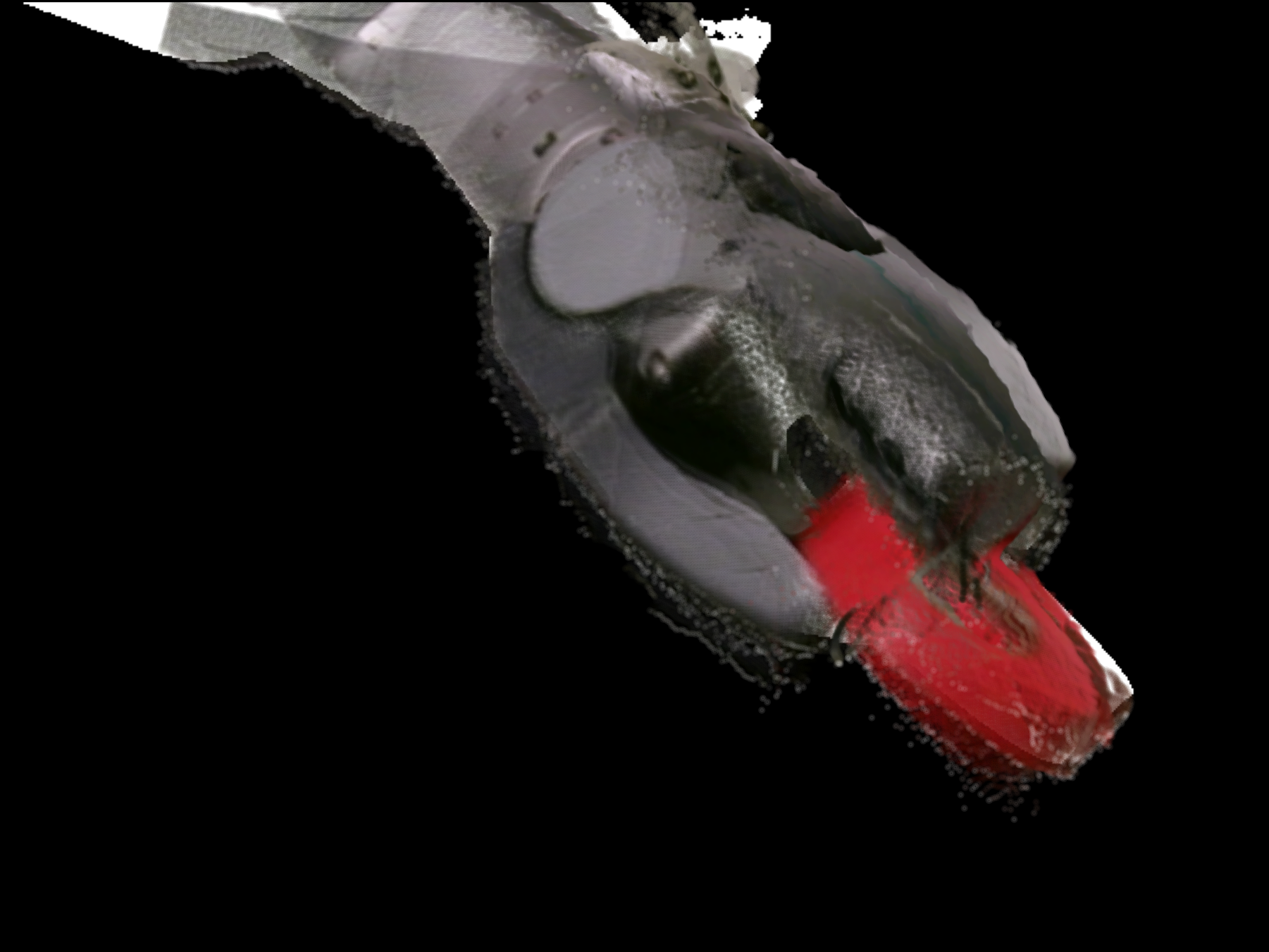}%
\caption{To evaluate the performance of GPE, we move the end-effector to test poses and retain ground truth images (\underline{left}). Then, we remove the background to extract a silhouette (\underline{center}). We project the reconstruction, transformed to the predicted test pose, to the camera's view and overlay it onto the silhouette (\underline{right}).}\label{fig:example_eval}
\end{figure}

\textbf{Metrics:} We wish to measure how closely, when the end-effector is moved to test pose, the reconstructed geometry matches the hidden ground truth images when the reconstruction is projected into the camera's view. An example evaluation is illustrated in \cref{fig:example_eval}. Note that both the estimated object poses and geometries affect how close matches are. We compute a symmetrised average minimum pixel distance between the pointset of our reconstruction and the pointset containing the ground truth silhouette of the gripper with the object, both projected to 2D camera view. Let us define the average minimum pixel distance $D$ between two sets of pixel coordinates $A$ and $B$ as
\begin{align}
D(A,B) = \frac{1}{\lvert A \lvert}\sum^{a\in A}\min_{b \in B} d(a, b),
\end{align}
where $\lvert A \lvert$ is the size of set $A$ and $d(\cdot,\cdot)$ denotes the $l2$ distance. As $D(A,B)\neq D(B, A)$, we symmetrise the distance via $\hat{D}(A,B)=0.5(D(A,B)+D(B,A))$
The ground truth silhouettes are extracted from captured images at the test end-effector poses. 

\textbf{Baselines: } A key contribution of this work is formulating the coordinate alignment problem from the outputs of the DUSt3R foundation model. Our formulation is \emph{structured}, in that we derive our estimator, given in \cref{eqn:estimator}, by considering the sequence of transformations from the robot's base to the external camera. Additionally, the loss of our formulation relies on the distance between training data and modelled points \emph{rendered} to the camera's view. To evaluate the quality of solutions from solving the formulated problem we compare this to the following baselines:
\begin{itemize}
\item \textbf{Structured formulation, without rendering:} We retain the structured estimator, but instead of minimising distances in the field of view of the camera, we directly minimise distances on $\mathrm{SE}(3)$, between the predicted pose and the pose given by the foundation model. We use the distance of $\mathrm{SE}(3)$ implementation in \cite{bregier2021deepregression}.
\item \textbf{Unstructured direct regression:} Instead of representing the estimator via the structured equation, we estimate the mapping between end-effector and object pose via direct regression on $\mathrm{SE}(3)$ a neural network. Approaches with a similar idea to eschew geometric reasoning, and directly learn transformations for camera calibration from data, have been explored in \cite{Valassakis2021LearningEC}. 
\end{itemize}

\subsection{Generalisation Evaluation Results}\label{subsec:gen_results}
The quantitative results of our evaluation at test end-effector poses are tabulated in \cref{tab:tool_handling}. Our approach, which is both structured and minimises distances between rendered points, outperforms the compared baselines. In particular, we observe that directly using a neural network to regress onto the object's pose does not generalise well to the unseen test poses. We also observe that the structured estimator where parameter values are minimised with respect to rendered points outperforms the alternative. The approach that does not render onto the camera view weighs the influence of each pose in the training set equally.
On the other hand, our approach of minimising distances rendered to the camera view gives more weight to images where the in-hand object is closer to the camera, producing better generalisation performance. In particular, we observe a particularly high error using the rendering-free approach with the grasped screwdriver. This is due to increasing errors in the pose estimates when the camera is far from the object. These images are generally given less weight when minimising the distance after rendering. Additionally, in \cref{fig:results} illustrate projections of the dense reconstruction onto example test set silhouettes, produced by our method. Here, we observe for each object type the projections of the dense reconstructions match closely with the white silhouettes. This indicates both the accuracy of the predicted pose as well as the estimated reconstruction. Images of dense reconstructions of the evaluated objects are also shown in \cref{fig:recon}.

\subsection{Geometry and Pose Estimation Under Limited Data}\label{subsec:limited_data}
\begin{figure}[t]
\centering
\begin{subfigure}[b]{0.33\linewidth}
\includegraphics[width=\linewidth]{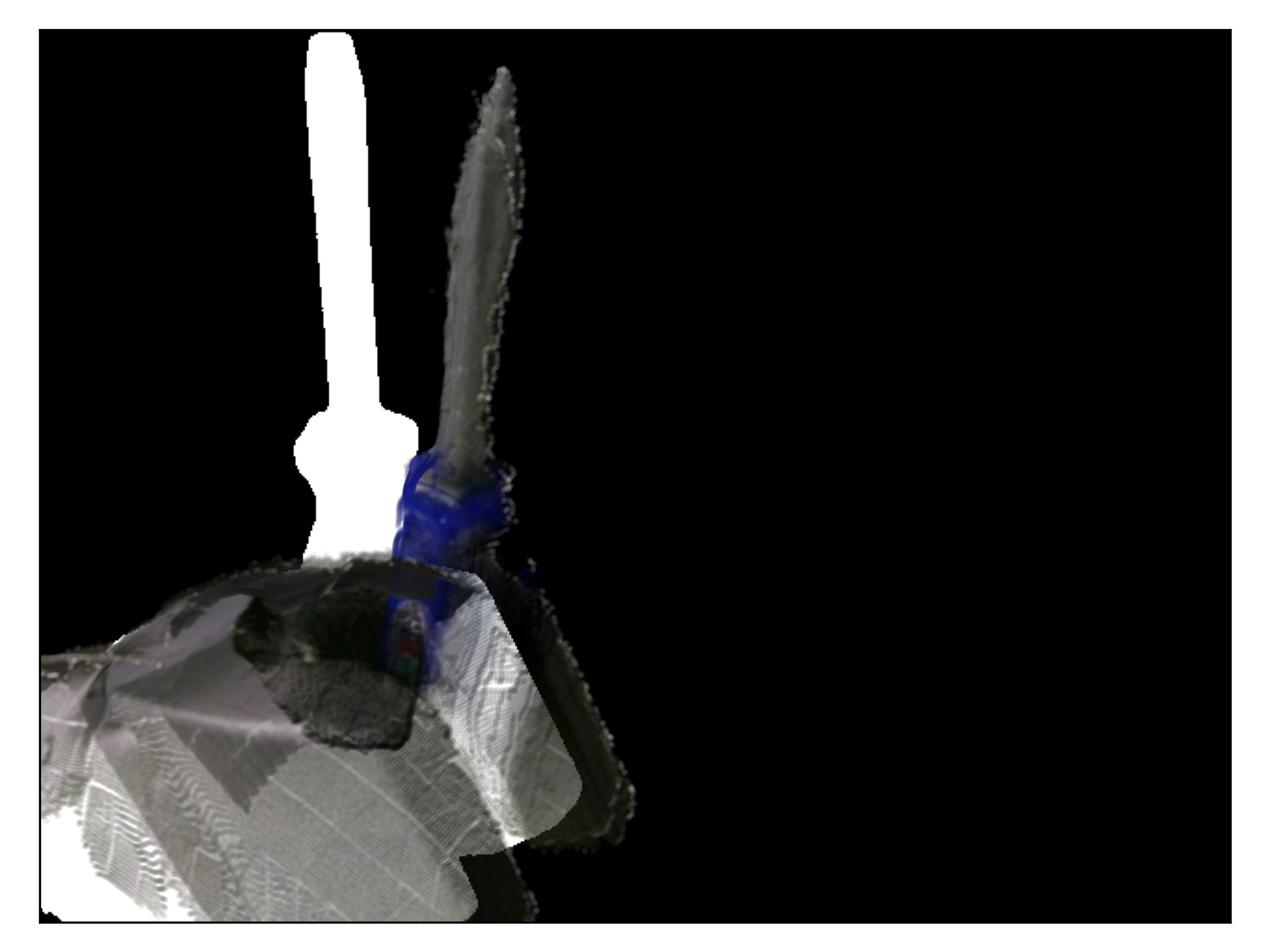}

\includegraphics[width=\linewidth]{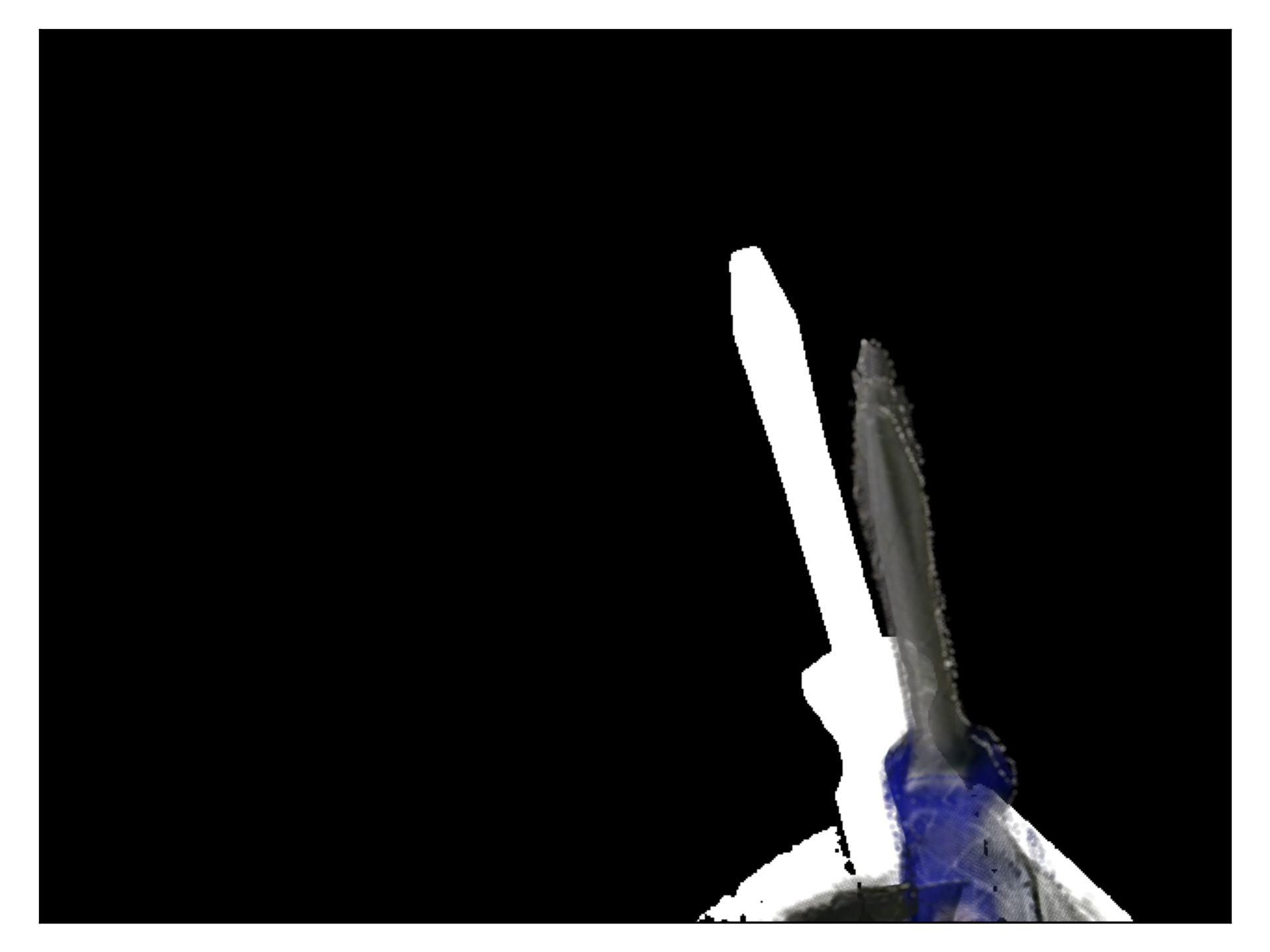}
\caption{3 input images}
\end{subfigure}%
\begin{subfigure}[b]{0.33\linewidth}
\includegraphics[width=\linewidth]{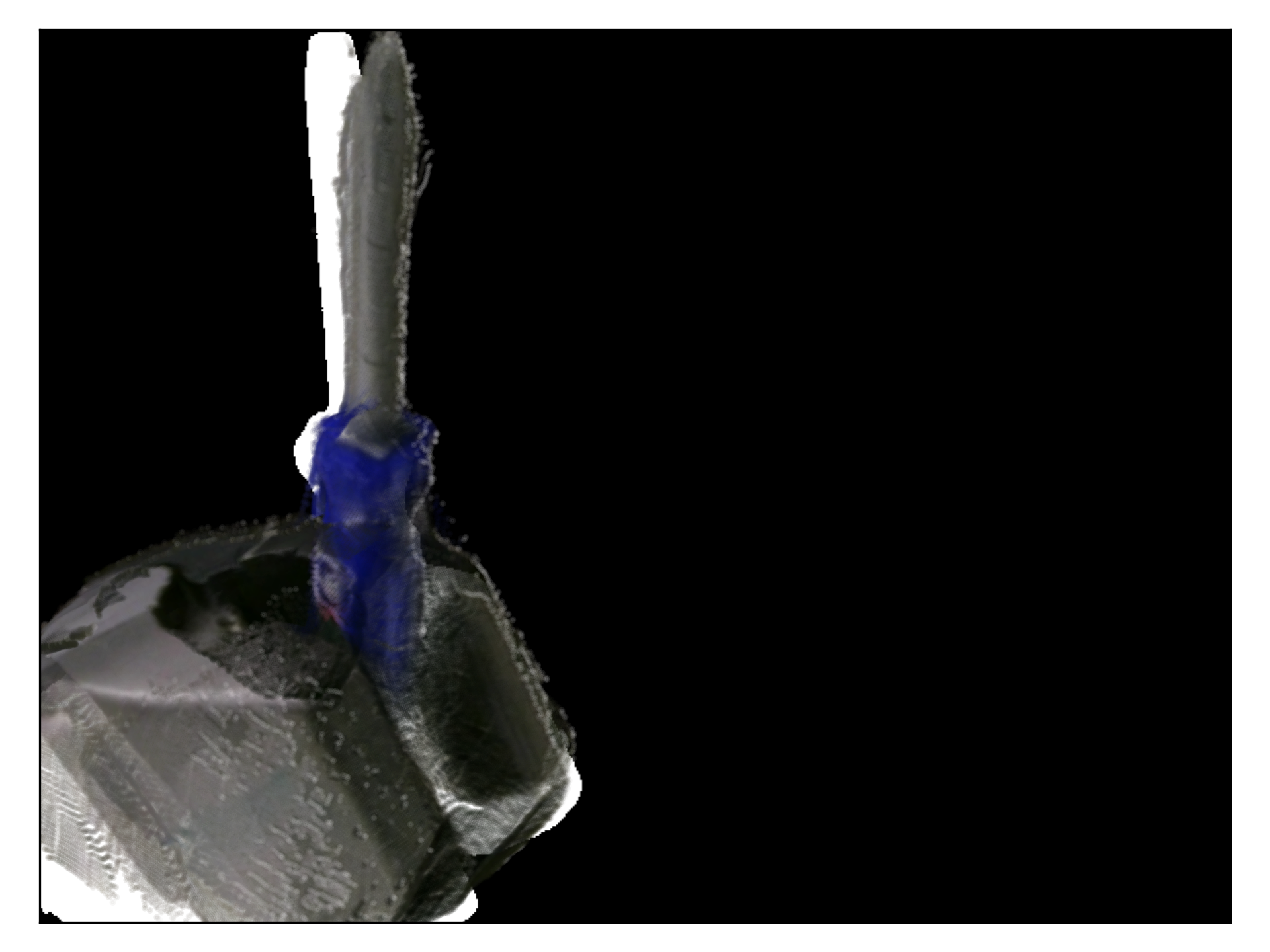}

\includegraphics[width=\linewidth]{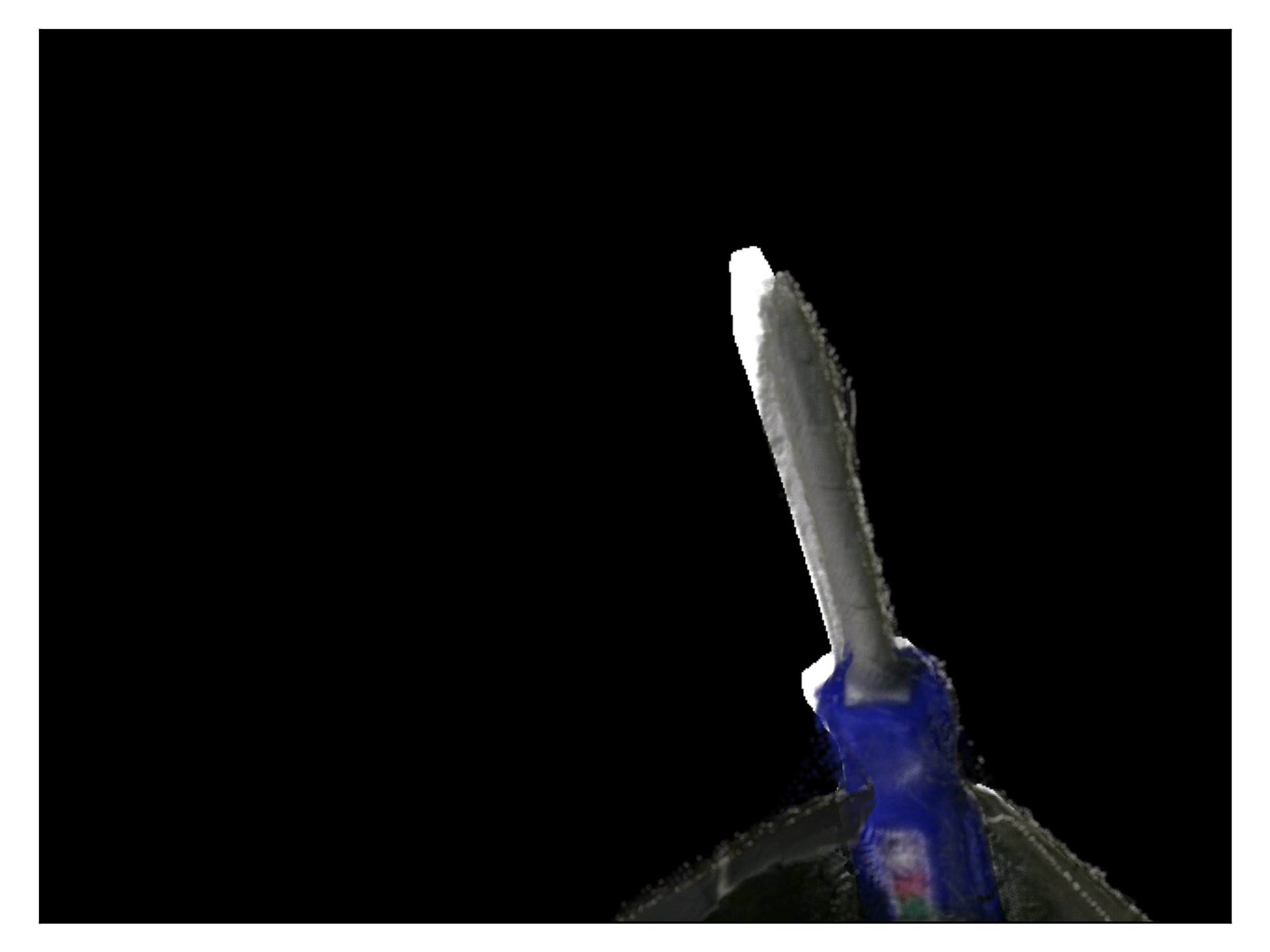}
\caption{6 input images}
\end{subfigure}%
\begin{subfigure}[b]{0.33\linewidth}
\includegraphics[width=\linewidth]{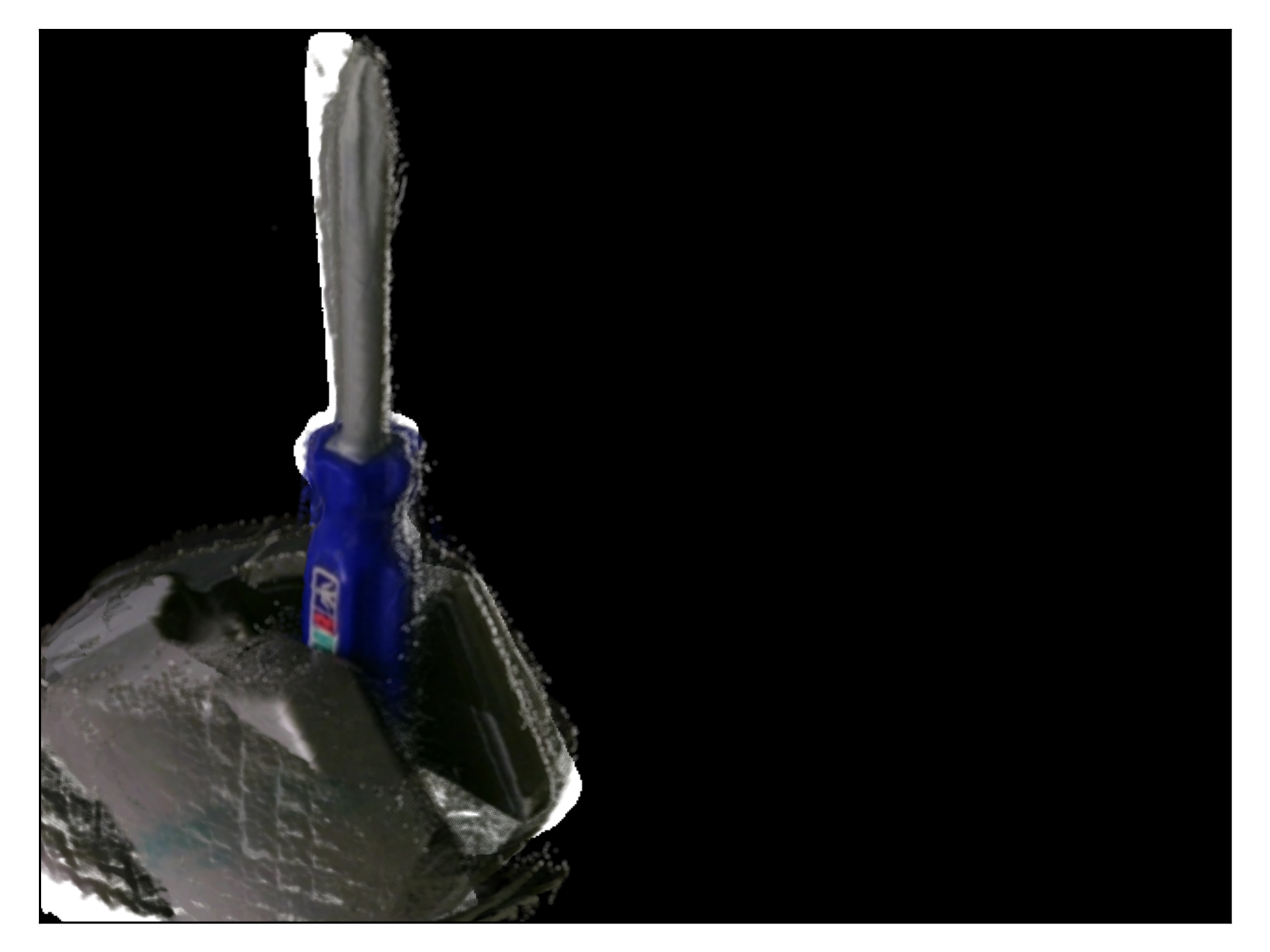}

\includegraphics[width=\linewidth]{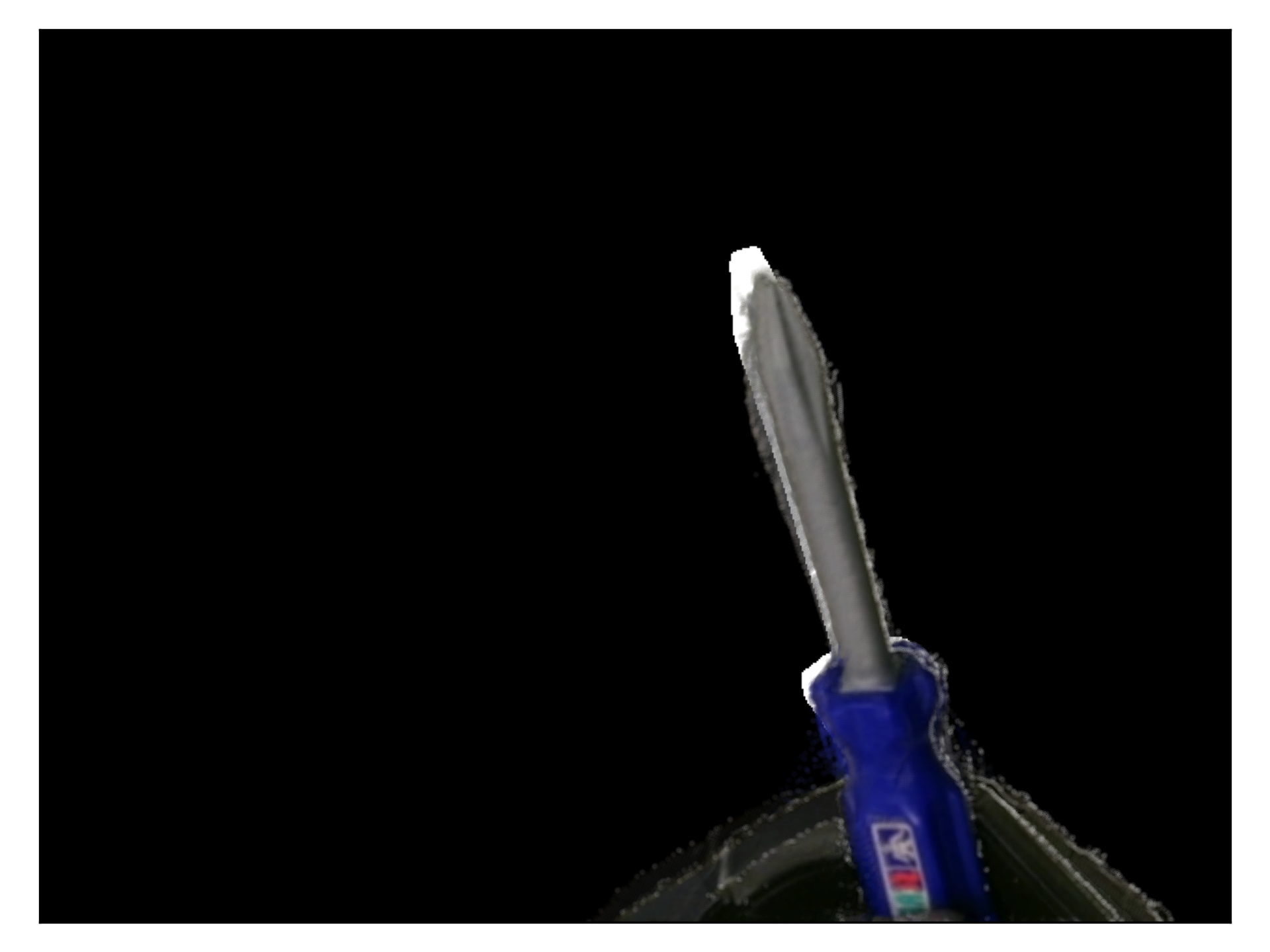}
\caption{9 input images}
\end{subfigure}
\caption{We investigate generalisation under limited training data (3,6 and 9 images for the left, middle and right columns respectively). Example test results of the reconstructed screwdriver, projected to camera view and laid over ground truth silhouettes, are shown. }\label{fig:input_imgs}
\end{figure}

\begin{table}[t]
\centering
\begin{adjustbox}{width=\linewidth}
\begin{tabular}{l|rrrrrr}
\toprule[1pt]\midrule[0.3pt]
         & \multicolumn{1}{l}{Hammer} & \multicolumn{1}{l}{Screwdriver} & \multicolumn{1}{l}{Wrench} & \multicolumn{1}{l}{Block} & \multicolumn{1}{l}{Tape} & \multicolumn{1}{l}{Brush} \\
         \midrule
3 Images & 34.90                      & 6.01                            & 5.24                       & 494.15                    & 46.32                    & 38.05                     \\
6 Images & 9.04                       & 1.90                            & 1.17                       & 0.62                      & 1.69                     & 0.98                      \\
9 Images & 1.42                       & 1.16                            & 1.11                       & 0.69                      & 0.94                     & 0.52\\                     
\midrule[0.3pt]\bottomrule[1pt]                      
\end{tabular}
\end{adjustbox}
\caption{We reduce the number of inputted images and report pixel distances. We observe that reducing the number of images to 6 has a limited impact on performance, while a large degradation occurs when the number is reduced to 3.}\label{table:decrease}
\end{table}

The bottleneck for generalisation under limited data lies in whether the coordinate-alignment problem can robustly generalise, even when the number of inputted images is small. To investigate this, we reduce the set inputted images from 9 down to 6 and 3. We report the pixel distances in \cref{table:decrease} and visualise in \cref{fig:input_imgs} example test images of the estimated in-hand screwdriver, for different numbers of inputted images. We observe that reducing the training data from 9 images to 6 images does not significantly affect generalisation, and is sufficient for tasks that do not require extreme precision. We take a closer look at the screwdriver example (\cref{fig:input_imgs}), with 6 input images, the tip of the rendered screwdriver remains close to the ground truth tip position. After a further decrease to 3 images, the performance is greatly reduced and the projections begin to visibly diverge from the silhouettes.

\begin{figure}[t]
\centering
\fbox{\includegraphics[width=0.485\linewidth]{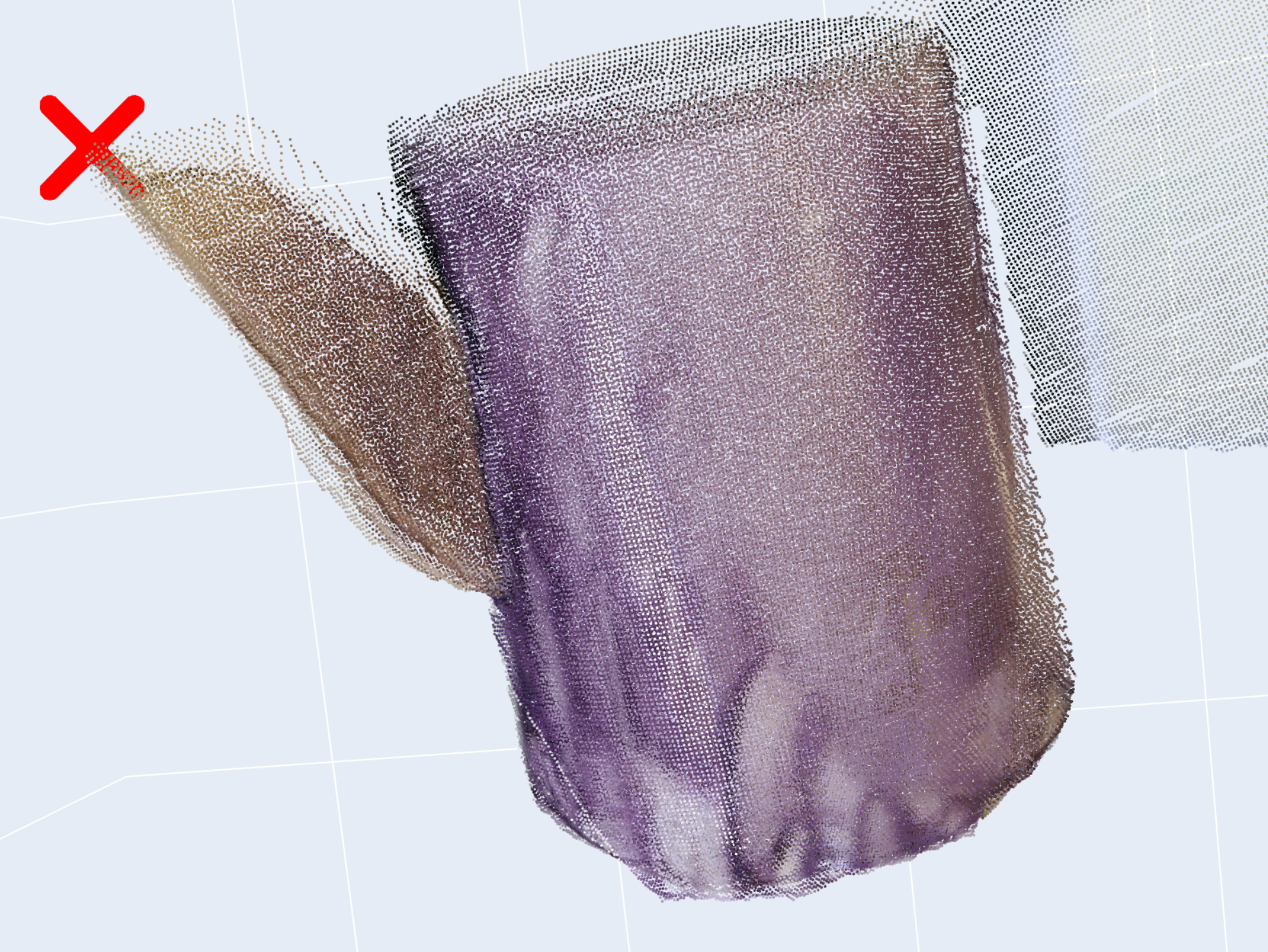}}%
\fbox{\includegraphics[width=0.485\linewidth]{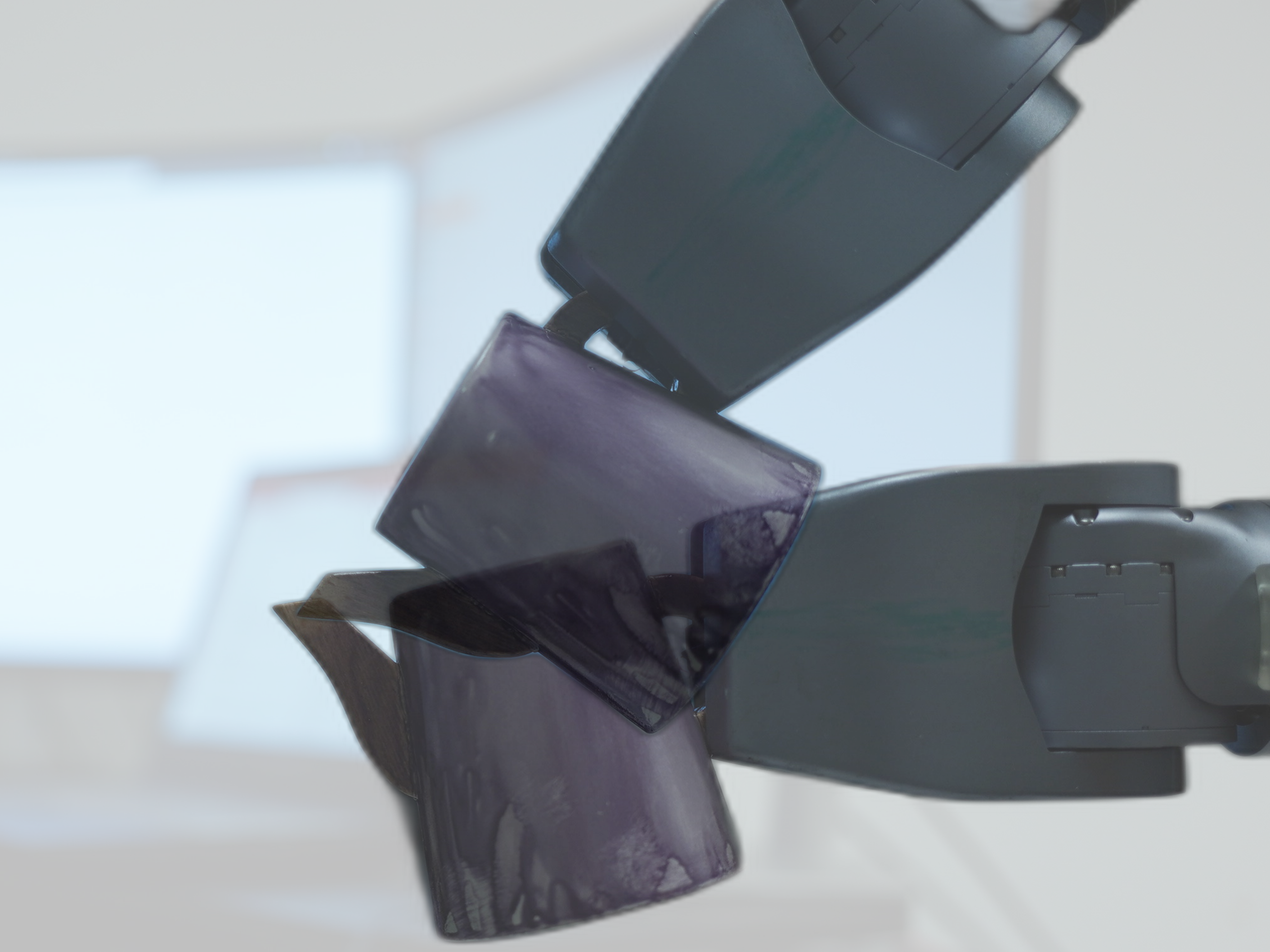}}%
\caption{We shape a pouring behaviour by specifying rotation at the tip of the estimated teapot. (Left:) The reconstruction of the estimated in-hand teapot. The coordinate on the teapot's spout marked by red cross is used as the pivot. (Right:) Overlapped images of before and after the pouring motion. We observe that translation at the tip of the pot is minimal, indicating an accurate estimate of the position of the spout's tip.}\label{fig:teapot}
\end{figure}

Here, we seek to shape the motion of pouring an in-hand teapot by tilting the pot down by $45$ degrees, with the tip of the teapot's spout as the pivot. We run GPE on six images taken by the external camera of the in-hand teapot. Then, we compute the goal configuration of rotating around the estimated pose of the spout tip via \cref{eqn:inv_mapping}. \Cref{fig:teapot} shows the held pot before and after the rotation around the pivot, taken by a fixed camera. We observe that translations of the tip of the spout are minimal. This indicates that both the geometry and the estimated end-effector-to-object transformation have been estimated accurately. These accurate estimates enable us to pinpoint the tip of the spout and subsequently use them to shape the robot's motion.

\subsection{Shaping Behaviour on Estimated Object}\label{subsec:shaping}
After estimating the geometry and the relative pose, we can shape the robot's behaviour by defining primitive motions relative to points on the held object. Here, we empirically demonstrate a simple example of using the outputs of GPE to design robot motions. 
\section{Conclusions and Future Work}\label{sec:conclusions}
In this paper, we tackle the problem of simultaneously estimating the geometry and the pose of objects held by a robot manipulator, from a small set of RGB images and end-effector poses. We introduce the joint Geometry and Pose Estimation (GPE) framework. GPE makes use of emerging 3D foundation models, trained for SfM tasks, to estimate both the object poses relative to the camera and object geometry. However, the outputs of these models are not in the coordinate frame of the robots and are not of accurate metric scale. We subsequently introduce the \emph{coordinate-alignment problem}. The solutions to this problem transform the reconstructed geometry into the robot's frame while scaling the outputs to be physically accurate. Kinematic equations mapping from robot joint angles to points on the object points can then be found. As a result, GPE provides us with the coordinates of points designated on the object, in the frame of the robot, as it moves its end-effector. We can then leverage this to design specific robot behaviour relative to points on the held object. We empirically show the robustness of GPE on a range of held objects, and under limited data. We also demonstrate that robot motion can be shaped relative to a point on a held object. 

Future work can revolve around developing more sophisticated motion planning and trajectory generation methods to better leverage the now-known geometry of the in-hand object. For example, these can now form a part of a collision-detector in sampling-based motion planners \cite{PDMP,rrts} or dynamical systems in structured policies \cite{GeoFab_gloabL_opt,Diff_templates,RMPs,Fast_diff_int}, or constraints and costs in trajectory optimisation \cite{CHOMP,Prob_struct_const,Kalakrishnan2011STOMPST}. Additionally, the small RGB images used are currently gathered by moving the end-effector to random coordinates in-view of the external camera. Subsequent research can also look at strategies to sequentially move to poses which actively maximise information gain. 
\bibliographystyle{ieeetr} 
\bibliography{bib}
\end{document}